\def\year{2021}\relax
\documentclass[letterpaper]{article} 
\usepackage{aaai21}  
\usepackage{times}  
\usepackage{helvet} 
\usepackage{courier}  
\usepackage[hyphens]{url}  
\usepackage{graphicx} 
\usepackage{natbib}  
\usepackage{caption} 
\usepackage{amsmath,amssymb, amsthm}
\usepackage{bm}
\usepackage{subcaption}
\usepackage[ruled, vlined, linesnumberedhidden]{algorithm2e}
\newtheorem{lemma}{Lemma}

\title{Learning Interpretable Models for Coupled Networks Under Domain Constraints }
\author{
    Hongyuan You, Sikun Lin, Ambuj K. Singh
    \\
}
\affiliations{
    Department of Computer Science \\
    University of California, Santa Barbara\\

    2120A Harold Frank Hall \\
    Santa Barbara, California 93106\\
    \{hyou, sikun, ambuj\}@ucsb.edu

}

\begin{document}

\maketitle

\begin{abstract}
Modeling the behavior of coupled networks is challenging due to their intricate dynamics. For example in neuroscience, it is of critical importance to understand the relationship between the functional neural processes and anatomical connectivities. Modern neuroimaging techniques allow us to separately measure functional connectivity through fMRI imaging and the underlying white matter wiring through diffusion imaging. Previous studies have shown that structural edges in brain networks improve the inference of functional edges and vice versa. In this paper, we investigate the idea of coupled networks through an optimization framework by focusing on interactions between structural edges and functional edges of brain networks. We consider both types of edges as observed instances of random variables that represent different underlying network processes. The proposed framework does not depend on Gaussian assumptions and achieves a more robust performance on general data compared with existing approaches. To incorporate existing domain knowledge into such studies, we propose a novel formulation to place hard network constraints on the noise term while estimating interactions. This not only leads to a cleaner way of applying network constraints but also provides a more scalable solution when network connectivity is sparse. We validate our method on multishell diffusion and task-evoked fMRI datasets from the Human Connectome Project, leading to both important insights on structural backbones that support various types of task activities as well as general solutions to the study of coupled networks.
\end{abstract}

\section{Introduction}

Recently, there has been an effort to move research from the investigation of single networks to the more realistic scenario of multiple coupled networks. In this paper, we consider the case of pairs of networks, $(\mathcal{G}_1, \mathcal{G}_2)$, that are categorized into different modalities over a population. Such coupled network systems can be found in infrastructures of modern society (energy-communication), financial systems (ownership-trade), or even human brains (anatomical substrate-cortical activation). Our goal is to reconstruct one network from information on the other, and during such a process obtaining a concise interpretation of how one network affects the other.

To achieve the above goal, we consider an edge-by-edge formulation. We treat one set of edges in $\mathcal{G}_1$ as predictors and the other set of edges in $\mathcal{G}_2$ as response variables in a multivariate linear regression model. Past research for the above problem relies on the restrictive Gaussian assumption, which simplifies the problem but is difficult to justify, especially in the domain of brain architectures. Adopting Gaussian assumption on non-Gaussian data can significantly prevent the detection of conditional dependencies and may lead to incorrectly inferred relationships among variables.

The learning of relationships between two different modalities can be difficult without sufficient data. As a result, in sparser data settings, the ability to specify constraints based on domain knowledge can be beneficial. For example, in the case of brain data, functional edges have mainly local influences, and structural edges are more responsible for long-distance influences \cite{rubinov2010complex,batista2018we}. We want preferences encoded in domain knowledge to guide the selection of partial correlations of unexplained noise terms in the constructed model. 

Based on the above motivations, we propose a flexible and efficient framework CC-MRCE (\underline{C}onvex-set \underline{C}onstrained \underline{M}ultivariate \underline{R}egression with \underline{C}ovariance \underline{E}stimation) that simultaneously learns both regression coefficients between two coupled networks and the correlation structure of noise terms. In a departure from existing methods, our framework encodes domain knowledge as a set of convex constraints and adopts a pseudolikelihood-based neighborhood-selection objective in partial correlation estimation, which has been shown to be more robust to non-Gaussian data. Because of the CC-MRCE objective's bi-convex nature, we alternately solve a regression sub-problem and a constrained partial correlation sub-problem until convergence. The latter sub-problem requires feasible solutions under given domain constraints that we render tractable via a modified two-stage proximal gradient descent method. 

We illustrate the use of our method in the context of the human brain. Brain data presents one of the greatest technical challenges in analysis and modeling due to a network-based characterization~\cite{bassett2017network, bassett2018nature}, non-Gaussian nature of data, high dimensionality, a small number of samples, and the need to incorporate domain knowledge. We apply the proposed framework on the Human Connectome Project (HCP) dataset~\cite{van2013wu}, where two coupled networks are constructed from fMRI scans (representing cortical activation) and diffusion scans (representing the anatomical substrate). We successfully predict a brain functional network from the given structural network; our method outperforms previous state-of-art methods, and our obtained models are easier to interpret. We investigate the structure-function coupling for seven different tasks. Our findings agree with the nature of fMRI task and brain region functions in existing literature, thus validating our model's ability to discover meaningful couplings.

Our main contributions are as follows:

\textbullet\ We propose a regularized multiple regression approach that adapts to non-Gaussian data. (Sections 2.1, 2.2)

\textbullet\ We incorporate prior domain knowledge to model estimation by formulating constraints into an optimization problem. (Section 2.3)

\textbullet\ We develop a fast method based on nested FISTA for solving the proposed optimization problem. (Section 3)

\textbullet\ We show the effectiveness of our model on HCP brain data using quantitative comparisons with existing approaches as well as a qualitative analysis. (Section 4.2)

\section{Problem Formulation}

In this section, we first introduce existing works on multiple regression under Gaussian assumptions and then motivate our approach under non-Gaussian settings and domain constraints.

\subsection{Multiple-output Regression Problem}
    
Let $\mathcal{D}$ be an $n$-subject sample set in which all subjects share the same coupling $(\mathcal{G}_1,\mathcal{G}_2)$ but have different edge values. For subject $i$ in $\mathcal{D}$, let $\bm{x}^{(i)}$ $=$  $(x_{1}^{(i)},\cdots, x_{p}^{(i)})$ be $p$-dimensional inputs that represent edge values in the first modality network $\mathcal{G}_1$, and $\bm{y}^{(i)} = (y_{1}^{(i)}, \cdots, y_{p}^{(i)})$ be $p$-dimensional outputs\footnote{In general, models do not require the same dimensions for inputs and outputs. We use the equality setting only for simplicity.} that stand for edge values in the second modality network $\mathcal{G}_2$. We assume that the inputs $\bm{x}_i$ and outputs $\bm{y}_i$ are correlated through a multivariate linear regression model:
    \begin{align}
    \label{linear-model-assumption}
        \bm{y}^{(i)} = \bm{x}^{(i)} \bm{B} + \bm{\epsilon}^{(i)}, \quad \text{for }i=1,...,n 
    \end{align}
    where $\bm{B}$ is the $p\times p$ regression coefficient matrix and its element $\beta_{jk}$ is the regression coefficient that measures the cross-modality impact of edge $x_{j}$ to edge $y_{k}$, and $\bm{\epsilon}^{(i)}$ is the noise vector of subject $i$. The model can be expressed in the matrix form:
    \begin{small}
    \begin{align}
    \label{linear-model-assumption-marix-form}
        \bm{Y} = \bm{X} \bm{B} + \bm{E}
    \end{align}
    \end{small}where row $i$ of $\bm{X} \in \mathbb{R}^{n\times p}$ and $\bm{Y} \in \mathbb{R}^{n\times p}$ are the structural and functional edge vectors $\bm{x}^{(i)}$ and $\bm{y}^{(i)}$ of subject $i$.
    
    A straightforward approach to estimating $\bm{B}$ is to solve $p$ separate regression problems, assuming noise terms are independent and uncorrelated. Recently, advanced methods have been proposed to exploit the correlation in noise terms to improve the modeling. They accomplish the goal by introducing an assumption that noise terms $\bm{\epsilon}^{(1)}$,...,$\bm{\epsilon}^{(n)}$ are all {\it i.i.d.} Gaussian $\mathcal{N}(\bm{0}, \bm{\Omega}^{-1})$ and then simultaneously estimating regression coefficients $\bm{B}$ and inverse covariance matrix $\bm{\Omega}$ of the noise terms. Two popular methods along this direction are MRCE  \cite{rothman2010sparse} and CGGM  \cite{sohn2012joint,wytock2013sparse,yuan2014partial}.  
    The MRCE method considers the conditional distribution $\bm{Y} \vert \bm{X} \sim \mathcal{N}(\bm{XB}, \bm{\Omega}^{-1})$ and estimates both $\bm{B}$ and $\bm{\Omega}$ by alternately minimizing the negative conditional Gaussian likelihood, with the $\ell_1$ lasso penalty applied on the entries of $\bm{B}$ and $\bm{\Omega}$. The other method, CGGM, further assumes that $\bm{X}$ and $\bm{Y}$ are jointly Gaussian. Under such formulation, the conditional distribution of $\bm{Y}\vert \bm{X}$ is given by $\mathcal{N}(-\bm{X} \bm{\Omega}_{XY} \bm{\Omega}^{-1}, \bm{\Omega}^{-1})$, which reparameterizes the regression coefficient $\bm{B}$ as $-\bm{\Omega}_{XY}\bm{\Omega}^{-1}$. Compared with MRCE, the objective of CGGM is based on the negative conditional Gaussian likelihood as well, but is jointly convex for $\bm{\Omega}_{XY}$ and $\bm{\Omega}$, and therefore more friendly to computation.

\subsection{Relaxing Gaussian Assumptions}

     Although MRCE and CGGM have received significant attention in solving multi-output regression problems, one drawback of these two approaches is the Gaussian assumption, especially for applications to brain data~\cite{freyer2009bistability, hlinka2011functional, eklund2016cluster}. Recall that MRCE assumes the Gaussian noise, and CGGM further assumes joint Gaussian distribution over both inputs and outputs. 
     We tested whether the HCP structural and functional data is Gaussian with a significance level of 0.05. The test rejects the Gaussian null hypothesis for 97.5\% of structural edges and 36.3\% of functional ones. Since our sample size is small, false negatives are more likely to occur \cite{columb2016statistical}, namely failing to reject the Gaussian hypothesis when the underlying data is non-Gaussian. Therefore, the proportion of non-Gaussian data in brain networks is expected to be even higher. Thus, relying on Gaussian assumptions is likely to affect the constructed models negatively.
     
     \noindent To avoid a Gaussian assumption, we propose a pseudolikelihood approach for learning multi-output regression models by optimizing the following objective function:
     \begin{small}
        \begin{align}
        \label{concord-mrce-objective-entrywise}
            &\mathop{\min}_{\lbrace\bm{B}_k\rbrace,\lbrace \omega_{jk}\rbrace}~
             \bigg[
            -n \sum_{j = 1}^{p} \log \omega_{jj} 
            +  \frac{1}{2} \sum_{j=1}^{p} \sum_{i=1}^{n} \bigg(  \omega_{jj} (\bm{y}_j^{(i)}-\bm{x}^{(i)}\bm{B}_j) \nonumber \\ 
            &+ \sum_{k \neq j} \omega_{jk} (\bm{y}_k^{(i)}-\bm{x}^{(i)}\bm{B}_k) \bigg)^2 
            + \lambda_1 \sum_{j < k}\vert \omega_{jk} \vert 
            + \lambda_2 \sum_{j < k} \vert \beta_{jk} \vert \bigg]
            \nonumber 
        \end{align}
    \end{small}
    or in a neat matrix notion:
    \begin{small}
    \begin{align}
        \mathop{\min}_{\bm{B},\bm{\Omega}}~ 
         -n\log\vert\bm{\Omega}_D\vert 
        & + \frac{1}{2}\text{tr}\left( (\bm{Y}-\bm{XB})^T(\bm{Y}-\bm{XB})\bm{\Omega}^2\right) \nonumber \\
        & + \lambda_1 \|\bm{B}\|_1 + \lambda_2 \|\bm{\Omega}_X\|_1
    \end{align}
    \end{small}where $\bm{\Omega}=\lbrace \omega_{jk} \rbrace$ denotes the inverse covariance matrix, $\bm{B}=\lbrace \beta_{jk} \rbrace$ denotes the coefficient matrix, and $\bm{\Omega}_D$ and $\bm{\Omega}_X$ denote the diagonal and off-diagonal parts of $\bm{\Omega}$. The proposed objective can be considered as a reparameterization of the Gaussian likelihood with $\bm{\Omega}^2$ and an approximation to the log-determinant term. It has been proven that under mild singularity conditions, such reparameterization can guarantee estimation consistency for distributions with sub-Gaussian tails \cite{peng2009partial, khare2015convex}.
    
    \par
    Next, we develop an optimization algorithm to minimize the objective. The objective function itself is not jointly convex for both variables $\bm{B}$ and $\bm{X}$, but remains convex with respect to each of them while keeping the other fixed. Therefore, we adopt the alternating minimization idea. In the $t$-th iteration, we first fix $\bm{B}$ as the estimated $\hat{\bm{B}}^{(t-1)}$ from the previous $(t-1)$-th iteration, and calculate the empirical covariance matrix $\bm{S}$ of noise terms:\begin{small}
    \begin{align}
        \label{concord-mrce-samplecov}
        \bm{S}^{(t-1)} = \frac{1}{n}(\bm{Y}-\bm{XB}^{(t-1)})^T(\bm{Y}-\bm{XB}^{(t-1)})
    \end{align}
    \end{small}Next, we estimate the inverse covariance matrix: $\bm{\Omega}^{(t)}$ with the given $\bm{S}^{(t-1)}$ as a constant:
    \begin{small}
    \begin{align}
        \label{concord-mrce-EM-omega}
        \bm{\Omega}^{(t)} = \mathop{\arg\min}_{\bm{\Omega}}
        &-\log\vert\bm{\Omega_D}\vert 
        + \frac{1}{2}\text{tr}\left(\bm{S}^{(t-1)}\bm{\Omega}^2\right)
        + \lambda_2 \|\bm{\Omega}_X\|_1
    \end{align}
    \end{small}
    Observe that the above subproblem follows CONCORD's original form, which is more robust to heavy-tailed data \cite{khare2015convex, koanantakool2017communication} than the conventional Gaussian likelihood approach and can be efficiently solved using proximal gradient methods with a convergence rate of ${\small{O(1/t^2)}}$ \cite{oh2014optimization}. Lastly, we keep $\bm{\Omega}$ fixed at ${\bm{\Omega}^{(t)}}^{2}$ and optimize the regression coefficients $\bm{B}$:
    \begin{small}
    \begin{align}
    \label{concord-mrce-EM-beta}
    \bm{B}^{(t)} = \mathop{\arg\min}_{\bm{B}}~  
        &\frac{1}{2}\text{tr}\left((\bm{Y}-\bm{XB})^T(\bm{Y}-\bm{XB})\bm{\Omega}^{(t)^2}\right) + \lambda_1 \|\bm{B}\|_1 
    \end{align}
    \end{small}
    Note that subproblem (\ref{concord-mrce-EM-beta}) is convex when ${\bm{\Omega}^{(t)}}^{2}$ is positive semi-definite. We present the above regression-based approach as CONCORD-MRCE in {\bf Algorithm 1} (pseudocode, see appendix).
    
\subsection{Imposing Domain Constraints}

Due to limited sample size of real-world datasets and their high dimensionality, incorporating accurate domain constraints can reduce the search space and avoid over-fitting. 

Under a linear mapping assumption, the partial correlation of response variables arises only from correlations in the noise terms. Therefore, $\bm{\Omega}$ not only represents the inverse covariance of noise terms, but also equals the conditional inverse covariance of $\bm{Y}\vert \bm{X}$. The nonzero entries of $\bm{\Omega}$ encode direct relationships among the target modality outputs $\bm{Y}$ that cannot be explained by weighted inputs $\bm{XB}$ of the source modality. It will be beneficial if the zero-vs-nonzero structure of $\bm{\Omega}$ is partially given by domain experts and used as hard constraints in model estimation. 

More formally, let $M$ be a binary matrix that has the same dimensions as $\bm{\Omega}$. We can define a convex matrix set $\mathbb{S}_M$, containing all matrices that share the same set of zero entries with $M$. We can then improve the previous regression-based approach to estimate $\bm{\Omega}$ under the domain constraint that takes the form of $\bm{\Omega} \in \mathbb{S}_M$, written in an equivalent unconstrained convex form:
\begin{small} 
\begin{align}
\label{NC-CONCORD}
\hat{\Omega} = \mathop{\arg\min}_{\bm{B},\bm{\Omega}}   
&  - \frac{n}{2} \log \vert \bm{\Omega}_D^2 \vert + \frac{1}{2} tr((\bm{Y}-\bm{X}\bm{B})^T (\bm{Y}-\bm{B}\bm{X})\bm{\Omega}^2)
\nonumber \\
& +\lambda_1 \| \bm{B} \|_1 + \lambda_2 \| \bm{\Omega}_X \|_1 + \mathbb{I}\lbrace \bm{\Omega} \in \mathbb{S}_M \rbrace
\end{align}
\end{small}where $\mathbb{I}\lbrace \bm{\Omega} \in \mathbb{S}_M \rbrace$ is an indicator function. This formulation can be extended to $\bm{\Omega} \in C$ whenever $C$ is a closed convex set of positive definite matrices.

\section{Solution}

In this section, we show how to adapt the previous solution when the inverse covariance $\bm{\Omega}$ is constrained during estimation. Notice that the ideas of {\bf Algorithm 1} can be mostly used to solve Eq (\ref{NC-CONCORD}), except for the $\bm{\Omega}$-update step, which is now affected by the added constraints. The new $\bm{\Omega}$-update step needs to solve the following sub-problem:
\begin{align}
        \label{cc-mrce-omega}
        \bm{\Omega}^{(t)} = \mathop{\arg\min}_{\bm{\Omega}}
        &-\log\vert\bm{\Omega}^2_D\vert 
        + \text{tr}\left(\bm{S}^{(t-1)}\bm{\Omega}^2\right) 
        \nonumber \\
        &+ \lambda_2 \|\bm{\Omega}_X\|_1
        + \mathbb{I}\lbrace \bm{\Omega} \in \mathbb{S}_M \rbrace
\end{align}
We follow the FISTA (\underline{F}ast \underline{I}terative \underline{S}oft-\underline{T}hresholding \underline{A}lgorithm \cite{beck2009fast}) approach that is used in CONCORD  \cite{oh2014optimization}. This method utilizes an accelerated gradient algorithm using soft-thresholding as its proximal operator for the $L_1$ norm and achieves a fast $O(1/t^2)$ convergence rate. Previous work~\cite{oh2014optimization} has also applied FISTA for partial correlation estimation and proved its efficiency. To adapt our constrained problem into the FISTA framework, we split our objective function (\ref{cc-mrce-omega}) into a smooth part and a non-smooth part:
\begin{align}
h_1(\bm{\Omega}) & = -\log \vert \bm{\Omega}_D^2 \vert + tr(\bm{S}\bm{\Omega}^2)  \nonumber \\
h_2(\bm{\Omega}) & = \lambda_2 \| \bm{\Omega}_X \|_1 +  \mathbb{I}\lbrace \bm{\Omega} \in \mathbb{S}_M \rbrace \nonumber
\end{align}
For any symmetric matrix $\bm{\Omega}$, the gradient of the smooth function can be easily calculated as:
\(\nabla h_1(\bm{\Omega}) = -2 \bm{\Omega}_D^{-1} + 2 \bm{\Omega} \bm{S}.\)
With this formulation, we now adapt the FISTA iterative scheme to solve our network-constrained problem (\ref{cc-mrce-omega}):
\begin{small}
\begin{align}
    \label{cFISTA-0}
    \alpha_{t+1} &= (1+\sqrt{1+4\alpha_t^2})/2 \\
    \label{cFISTA-1}
    \bm{\Theta}^{(t+1)} & = \bm{\Omega}^{(t)} + \frac{\alpha_{t}-1}{\alpha_{t+1}}(\bm{\Omega}^{(t)} - \bm{\Omega}^{(t-1)}) \\
    \bm{\Omega}^{(t+1)} & = \operatorname{prox}_{\gamma h_2}[\bm{\Theta}^{(t+1)} - (n\tau_t/2) \nabla h_1(\bm{\Theta}^{(t+1)})] \label{cFISTA-2}
\end{align}
\end{small}where $\tau_{t}$ is the step length and $t$ denotes the iteration number. $\gamma$ is a trade-off parameter that controls the extent to which the proximal operator maps points towards the minimum of $h_2(\bm{\Omega})$, with larger values of $\gamma$ associated with larger movement near the minimum.

In these iterative steps, $\bm{\Theta}_{t+1}$ is an expected position, updated purely by momentum. Within each loop, the algorithm first takes a gradient step of the estimated future position (Eq.\ref{cFISTA-1}) and then applies the proximal mapping of a closed convex function $h_2(\bm{\Omega})$.

In contrast to the standard FISTA approach, the composite function $h_2(\bm{\Omega})$ consists of a sparsity penalty and a network-constrained indicator function. More specifically, we can write down the explicit form of Eq.~(\ref{cFISTA-2}) according to the proximal operator definition:
\begin{small}
\begin{alignat}{2}
    &\hat{\bm{\Omega}} &&= \hat{\bm{\Omega}}_X + \bm{A}_D  \nonumber \\ 
    \label{cFISTA-primal}
    &\hat{\bm{\Omega}}_X 
    &&= \operatorname{prox}_{\gamma h_2}(\bm{A}_X) \nonumber \\
    & &&= \mathop{\arg\min}_{\bm{\Omega}_X \in C} \frac{1}{2\gamma}\| \bm{\Omega}_X - \bm{A}_X \|_{F}^2 
    + \lambda_2\|\bm{\Omega}_X \|_1  
\end{alignat} 
\end{small}where $\bm{A} = \bm{\Theta}^{(t+1)} - (n\tau_t/2)\nabla h(\bm{\Theta}^{(t+1)})$. Instead of directly solving the original problem (\ref{cFISTA-primal}), we consider its dual problem as follows. Let matrix $\bm{H}$ be the dual variable of matrix $\bm{\Omega}$. We have:
\begin{small}
\begin{align}
    & \min_{\bm{\Omega}_X \in C} ~ ( \frac{1}{2\gamma} \| \bm{\Omega}_X - \bm{A}_X \|_F^2 + \lambda_2  \max_{\|\bm{H}_X\|_{\infty} \leq 1} \text{vec}(\bm{H}_X)^T \text{vec}(\bm{\Omega}_X) ) \nonumber \\
    = & \max_{\|\bm{H}_X\|_{\infty} \leq 1}  \min_{\bm{\Omega}_X \in C}~
        \frac{1}{2\gamma} \Big( \| \bm{\Omega}_X - (\bm{A}_X - \gamma \lambda_2 \bm{H}_X) \|_F^2 \nonumber \\
    & \hspace{6em} - \| \bm{A}_X - \gamma \lambda_2 \bm{H}_X \|_F^2 + \|\bm{A}_X\|_2^2 \Big) \label{dual-problem}
\end{align}
\end{small}where $\bm{A}_D$ and $\bm{A}_X$ denote the diagonal and off-diagonal part of $\bm{A}$. Since the initial objective function above is convex in $\bm{\Omega}_X$ and concave in $\bm{H}_X$, we exchange the order of the minimum and maximum operator in which the inner minimization problem has an obvious solution through orthogonal projection theorem \cite{rockafellar1970convex}, written as 
\begin{small}
\begin{align}
\label{cFISTA-dual-primal-correspondence}
\bm{\Omega}_X  = \mathbb{P}_C~(\bm{A}_X - \gamma \lambda_2 \bm{H}_X)
\end{align}
\end{small}where $\mathbb{P}_{C}$ is defined as an projection operator: 
$ \mathbb{P}_{C}(\bm{\Gamma}) =\operatorname{argmin}_{\bm{R} \in C} \|\bm{R} - \bm{\Gamma}\|_F^2$
and its orthogonal projection operator $\mathbb{P}_{C^{\perp}}$ is defined as $I - \mathbb{P}_{C}$. In the special case that $C = \mathbb{S}_M$, projection $\mathbb{P}_{C}(\bm{\Gamma})$ is equivalent to removing invalid nonzero entries of the input matrix $\bm{\Gamma}$. 

Inserting the optimal $\bm{\Omega}_X$ back into objective (\ref{dual-problem}), we now obtain the final dual form of the problem (\ref{cFISTA-primal}):
\begin{small}
\begin{align}
    \label{BT-1}
    \hat{\bm{H}}_X 
    =  \mathop{\arg\min}_{\|\bm{H}_X\|_\infty \leq 1}
    &\| \bm{A}_X - \gamma \lambda_2 \bm{H}_X \|_2^2 \nonumber \\ 
    &- \| \mathbb{P}_{C^{\perp}}(\bm{A}_X-\gamma \lambda_2 \bm{H}_X) \|_2^2
\end{align}
\end{small}where any solution $\hat{\bm{H}}_X$ to the dual problem corresponds to a primal solution through Eq.(\ref{cFISTA-dual-primal-correspondence}).  Since the dual objective is continuously differentiable and constraints on $l_{\infty}$-norm are convex, we can again efficiently solve it with additional inner FISTA iterations, which is to minimize an equivalent composite objective $\min g_1(\bm{H}_X) + g_2(\bm{H}_X)$, where $g_1(\bm{H}_X)$ is smooth and $g_2(\bm{H}_X)$ is non-smooth:
\begin{small}
\begin{align}
g_1(\bm{H}_X) & = \| \bm{A}_X -\gamma \lambda_2 \bm{H}_X \|_2^2 - \| \mathbb{P}_{C^{\perp}}(\bm{A}_X -\gamma \lambda_2 \bm{H}_X) \|_2^2  \nonumber \\
g_2(\bm{H}_X) & = I_{\lbrace \|\bm{H}_X\|_\infty \leq 1 \rbrace}(\bm{H}_X). \nonumber
\end{align}
\end{small}To adapt FISTA to the problem (\ref{BT-1}), the only thing left is to obtain the gradient of smooth function $g_1(\bm{H}_X)$, for which we need the lemma below:

\begin{lemma} \cite{moreau1965proximite} If $g$ is a closed proper convex function, and for any positive $t$, define a proximal operator $g_t(\bm{x}) := \operatorname{inf}_{\bm{u}} \left[g(\bm{u}) + \frac{1}{2t} \| \bm{u} - \bm{x} \|^2 \right]$, then its infimum is attained at the unique point $\operatorname{prox}_t(g)(\bm{x})$. Further, $g_t$ is continuously differentiable on $\mathbb{E}$ with a $1/t$-Lipschitz gradient given by $\nabla g_t(\bm{x}) = (\bm{x} - \operatorname{prox}_t(g)(\bm{x}))/ t$.
\end{lemma}

\noindent According to Lemma 1, we simply plug $g(\bm{u}) = \delta(\bm{u} \in C)$, and obtain that $\operatorname{prox}_t(g)(\bm{x}) = \mathbb{P}_C(\bm{x})$. Therefore the gradient $\nabla g_1$ is calculated as:
\begin{align}
    \label{g1}
    \nabla g_1(\bm{H}_X) = -2\gamma \lambda_2 \cdot \mathbb{P}_C(\bm{A}_X - \gamma \lambda_2 \bm{H}_X).
\end{align}
and the proximal mapping of function $g_2(\bm{\Omega})$ becomes a projection of $\bm{H}_X$ into the $L_\infty$-ball:
\begin{align}
    \left(\operatorname{prox}_{g_2}(\bm{H})\right)_{ij} 
    &= sign(\bm{H}_X) \min \lbrace \vert \bm{H}_X \vert, \textbf{1}_X \rbrace 
\end{align}

This completes the modified $\bm{\Omega}$-update step in the constrained setting (referred to as Constrained-CONCORD). Its corresponding pseudocode is presented in \textbf{Algorithm 3} (pseudocode, see appendix). The overall framework of simultaneously estimating $\bm{\Omega}$ and $\bm{B}$ is presented in \textbf{Algorithm 2} (pseudocode, see appendix), which we name as CC-MRCE. 

\section{Experiments}
\label{sec:exp}

We present two sets of experiments. First, to understand our method's strengths and limitations, we utilize simulated datasets that allow us to inspect both reconstruction performance and model selection performance. We compare the proposed method CC-MRCE with other baselines when the underlying data distribution does not follow the Gaussian assumption. We show that both the non-Gaussian assumption and the network constraints contribute to the improvement of performance. Second, we conduct experiments on the Human Connectome Project (HCP) data \cite{van2013wu}. Our model offers a quantitative advantage over baseline methods for predicting functional networks from structural ones; at the same time, our results agree with existing neuroscience literature.
    
    \subsection{Application to Simulated Data}
    
    \begin{table*}[htbp]
        \centering
        \begin{tabular}{l|c|ccc|cc} \hline
          Model  &
          \begin{tabular}[c]{@{}c@{}}Reconstruction \\ MSE percentage (100\%)\end{tabular} &
          \begin{tabular}[c]{@{}c@{}}Pearson's \\ $r$-score \end{tabular} &
          \begin{tabular}[c]{@{}c@{}}min\\ p-value\end{tabular} &
          \begin{tabular}[c]{@{}c@{}}max\\ p-value\end{tabular} &
          \begin{tabular}[c]{@{}c@{}}relative AUC \\ w.r.t. $\bm{\Omega}$ \end{tabular} &
          \begin{tabular}[c]{@{}c@{}}relative AUC \\ w.r.t. $\bm{B}$\end{tabular} 
          \\ \hline
            CC-MRCE (unconstrained) & 50.88  & 0.709 & 5.146E-09 &  0.101 & 0.520 & 0.536 \\ \hline
            CC-MRCE (SNR:1.0)   & 43.24     & 0.763 & 2.265E-11 &  0.083 & 0.855 & 0.640 \\ \hline
            CC-MRCE (SNR:2.0)   & 43.18     & 0.764 & 2.109E-11 &  0.073 & 0.925 & 0.662 \\ \hline
            CC-MRCE (perfect) & 42.98  & 0.764 & 2.692E-11 &  0.064 & 1.000 & 0.671 \\ \hline
            MRCE                & 60.22     & 0.653 & 8.686E-09 &  0.174 & 0.375 & 0.540 \\ \hline
            CGGM                & 76.21     & 0.552 & 2.141E-07 &  0.965 & 0.330 & 0.195 \\ \hline
        \end{tabular}
        \caption{Reconstruction performance and model selection performance of models on simulated dataset. CC-MRCE variations uniformly outperform MRCE and CGGM. CC-MRCE variations with more strict constraints perform better. }
        \label{table:syn-exp}
    \end{table*}

\textbf{Data Generation.} 
Using a similar approach to existing works \cite{peng2009partial,khare2015convex,rothman2010sparse}, we generate our simulated dataset by first synthesizing two key model parameter matrices $\bm{\Omega}_0$ and $\bm{B}_0$, and then construct input, output, and noise terms ({\it i.e.} $\bm{x}^{(i)}$, $\bm{y}^{(i)}$ and $\bm{\epsilon}^{(i)}$'s).
    
Note that every positive-definite matrix has a Cholesky decomposition that takes the form of $\bm{L} \bm{L}^{T}$, where $\bm{L}$ is a lower triangular matrix $\bm{L}$, and if $\bm{L}$ is sparse enough then $\bm{L} \bm{L}^{T}$ is sparse as well. Therefore, we first sample a sparse lower triangular $\bm{L}$ with real and positive diagonal entries, and then generate our inverse covariance matrix with $\bm{\Omega}_0 = \bm{L} \bm{L}^{T}$. The generated $p \times p$ positive definite matrix $\bm{\Omega}_0$ has $10\%$ nonzeros entries and a condition number of $4.3$. To demonstrate the robustness of proposed method CC-MRCE on non-Gaussian data, we sample the noise terms $\lbrace \bm{\epsilon}^{(i)} \rbrace_{i=1}^{n}$ according to a multivariate {\it t}-distribution with zero mean and covariance matrix $\bm{\Sigma} = \bm{\Omega}_0^{-1}$. 
    
Next, we generate a sparse coefficient matrix $\bm{B}_0$ using the matrix element-wise product trick, $\bm{B}_0 = \bm{W} \circ \bm{K} \circ \bm{Q}$. In this construction approach, $\bm{W}$ has entries with independent draws from standard normal distribution $\mathcal{N} (0, 1)$. Each entry in $\bm{K}$ is drawn independently from a Bernoulli distribution that takes value $1$ with probability $s_1$. $\bm{Q}$ has rows that are either all one or all zero, which are determined by independent Bernoulli draws with success probability $s_2$. Generating the sparse $\bm{B}_0$ in this manner, we not only control its sparsity level, but also forcibly make $(1-s_2)p$ predictors to be irrelevant for $p$ responses, and guarantee that each relevant predictor is associated with $s_1 p$ response variables. 
    
In the following experiments, the probabilities $s_1$ and $s_2$ are chosen to be $0.15$ and $0.8$, the sample size $n$ is fixed at $50$, and the input and output dimensions $p$ and $q$ are both set to $20$. The input $\bm{x}^{(i)}$'s are sampled from a multivariate normal distribution $\bm{\mathcal{N}}(0,\bm{\Sigma_{X}})$ where $(\bm{\Sigma_{X}})_{jk} = 0.5 ^ {\vert j - k \vert}$, following previous works \cite{yuan2007model,peng2010regularized}. The output $\bm{y}^{(i)}$'s are calculated as the linear model assumption in Eq.(\ref{linear-model-assumption}). We replicate the above process for independently generating a validation dataset of the same sample size. All penalty parameters are selected simultaneously and tuned according to the validation error.

\textbf{Methods.} In this heavy-tailed setting, we compare the performance of CC-MRCE to CGGM and MRCE, which are developed under Gaussian settings. The CGGM implementation that we used in this experiment is provided by \cite{mccarter2016large} and has been optimized for large-scale problems and limited memory. The MRCE implementation is provided by \cite{rothman2010sparse}. Both CGGM and MRCE do not adapt to the network constraints we impose here. 
    
In order to inspect the effectiveness of network constraints, we apply multiple variations of constraint sets to the proposed CC-MRCE method, designated as CC-MRCE (unconstrained), CC-MRCE (SNR: 2.0), CC-MRCE (SNR: 1.0) and CC-MRCE (perfect). {\it Network constraints} in each variation are defined as follows. For CC-MRCE (perfect), we choose $\mathbb{S}_E = \lbrace \bm{\Omega} : \bm{\Omega} (j,k)=0$ if $\bm{\Omega}_0 (j,k)=0 \rbrace$, which forces selected nonzeros in solution $\bm{\Omega}$ to completely fall into ground-truth nonzeros. For CC-MRCE (SNR: 2.0) and CC-MRCE (SNR: 1.0), we loosen the feasible set by adding $50\% \| \bm{\Omega}_0\|_0$ and $100\%\| \bm{\Omega}_0\|_0$ spurious nonzero positions, which are randomly sampled from positions of zero entries in $\bm{\Omega}_0$. For CC-MRCE (unconstrained), we remove all constraints so that the method only relies on regularized multi-regression. 
  
\begin{figure}[h]
    \centering
    \includegraphics[width=0.98\linewidth]{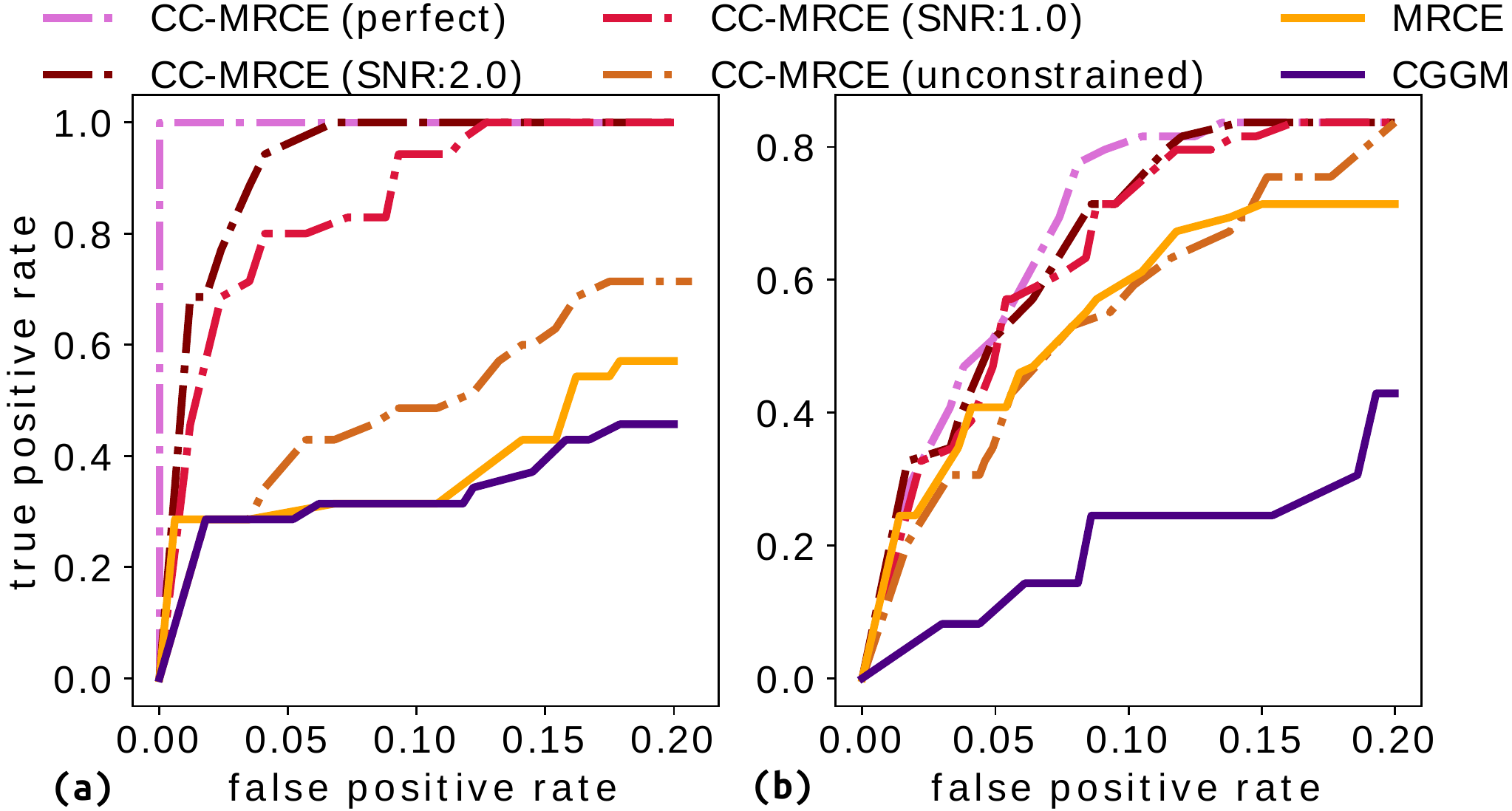}
    \caption[center]{(a) ROC curve for $\bm{\Omega}$ estimation, (b) ROC curve for $\bm{B}$ estimation. Benefit from domain constraints, CC-MRCE obtains better ROC curves when uncovering nonzero entries of $\bm{B}$ and $\bm{\Omega}$.}
    \label{fig:roc_curve}
\end{figure}

\textbf{Performance Evaluation.} We evaluate the reconstruction performance of models using the conventional MSE error (in percentage).  Correlation coefficients between predicted and ground-truth outputs are also provided along with their corresponding p-values. In addition, we use the relative area under the curve (AUC) of receiver operating characteristic (ROC) curves \cite{fawcett2006introduction, friedman2010applications}, with regards to $\bm{\Omega}$ and $\bm{B}$, as key measures to compare model selection performance of all these methods. The AUC of a perfect ROC curve, which would be 1, indicates an ideal recovery of gound truth zero-vs-nonzero structure in $\bm{\Omega}$ (or $\bm{B}$). However, models with large false-positive rates (FPR) are barely meaningful in real scenarios. To focus on the initial portion of ROC curves, we control the FPRs simultaneously to be smaller than $0.2$ for both $\bm{\Omega}$ and $\bm{B}$ estimation. Thus, the maximum AUC value that a model can reach is just $0.2$. For ease of comparison, we provide relative AUC values, divided by $0.2$ to normalize to $1$. For each method, we run the algorithm with at least $25$ appropriate parameter pairs $(\lambda_1, \lambda_2)$ to get its ROC curve. Recall that all methods in this section are required to estimate $800$ parameters given $n=50$ samples.
    
\textbf{Test Results.} Table \ref{table:syn-exp} displays the test results of six different settings in all measures mentioned above. As can be seen, all four CC-MRCE variations (with different network constraints) obtain significantly smaller reconstruction MSE percentages, higher correlation coefficients, and smaller $p$-values. Note that CGGM behaves the worst in all measures since it is deeply rooted in the Gaussian setting and is consequently misled by these assumptions. We also see that the performance of CC-MRCE gradually improves when the applied network constraints are more informative.  
    
We also plot ROC curves for $\bm{\Omega}$ and $\bm{B}$  estimation in Figure \ref{fig:roc_curve}.(a) and Figure \ref{fig:roc_curve}.(b), respectively. It is clear that CC-MRCE performs better than MRCE and CGGM, across different choices of network constraints. For $\bm{\Omega}$ estimation, as expected, CC-MRCE with more strict constraints has steeper curves, suggesting that it recovers mostly correct partial relationships between variables with very few spurious connections, and therefore achieves higher AUC scores. CC-MRCE (perfect) behaves perfectly for $\bm{\Omega}$-ROC, by its definition. For the estimation of matrix $\bm{B}$, a similar phenomena demonstrates that network constraints improve the learning of regression coefficients and lead to a better reconstruction performance. Without imposing network constraints, unconstrained formulation of CC-MRCE is likely to generate a biased estimate of $\hat{\bm{\Omega}}$ on small datasets and can not recover ground truth features in $\bm{B}$.

\subsection{Application to Human Connectome Data}

\textbf{Problem Formulation.} Many works in literature report on coupling between brain structural connectivities (SCs) and functional connectivities (FCs) for both resting state \cite{hagmann2008mapping,hermundstad2013structural,honey2009predicting,honey2007network} and task-evoked states \cite{hermundstad2013structural,raichle2015brain,cole2014intrinsic,davison2015brain}. One such recent work \cite{becker2018spectral} maps SC to resting-state FC by aligning both the eigenvalues and eigenvectors of a subject's SC and FC matrices, and evaluates the mapping by reconstructing resting FCs from SCs. 

Inspired by the domain observation \cite{batista2018we} that two functional connections sharing the same node are more likely to form a meaningful pathway or functional activation pattern, we constrain the partial correlations between non neighboring edges to be zero, i.e. $\bm{\Omega}_{jk}=0$ if $e_j$ and $e_k$ are not incident edges in the fMRI-constructed network.

We conducted experiments by reconstructing task FCs from SCs of 51 subjects with their fMRI data from 
HCP dataset under seven task states.
We used the parcellation scheme in \cite{hagmann2008mapping} with a spatial scale of 33, resulting in 83 brain regions and 3403 possible edges.
Columns in predictor matrix, Equation (\ref{linear-model-assumption-marix-form}), $\bm{X} \in \mathbb{R}^{51 \times 3403}$ represent SCs, whose entries are numbers of white matter streamlines intersecting pairs of brain regions, and columns in response matrix $\bm{Y} \in \mathbb{R}^{51 \times 3403}$ represent FCs, whose entries are functional correlations between cortical activities of brain regions. $\bm{B}$ denotes the mapping (or coupling) between SCs and FCs, and $\bm{E}$ denotes the part of FCs that cannot be explained by SCs. We also normalized SC values to (0, 1], a range comparable to FC.
We ran a 10-fold cross-validation of our model for each task, splitting data in a 9-1 train-validation ratio (46-5 split for the 51 subjects). We selected the optimal hyperparameters \(\lambda_1\) and \(\lambda_2\) by a \(5\times 5\) grid search in the log-scale between \(10^{-1.6}\) to \(10^{-0.4}\), keeping the models with smallest Mean Squared Error (MSE) percentage
on the validation sets, averaged across 10 folds. In our case, both \(\lambda_1\) and \(\lambda_2\) have optimal values around \(0.1\) across tasks. Aside from MSE, we also tested the Pearson correlation coefficient between predicted FCs and ground truth FCs, (referred to as Pearson's $r$-score, listed in Table \ref{table:task-exp}) and minimum, maximum $p$-values. The reconstruction MSE percentage is below 1\% for the training data and around 8\% for the validation data. Strong and significant positive correlations are shown for both training ($r$-score around 0.6 to 0.8) and validation ($r$-score around 0.5) data. 
These results indicate our model's effectiveness in FC reconstruction by exploiting cross-modal coupling between SC-FC and domain prior knowledge on FC-FC relationships.

\begin{table}[t]
\footnotesize
\centering
\begin{tabular}{c|lcc} \hline
Tasks & &
  \begin{tabular}[c]{@{}c@{}}Reconstruction \\ MSE percentage \\ (100\%)\end{tabular} &
  \begin{tabular}[c]{@{}c@{}}Pearson's \\ $r$-score\end{tabular}\\ \hline
EMOTION    & 1              & 89.78 $\pm$ 21.95   & -0.0309 $\pm$ 0.0285 \\
          & 2               & 81.57 $\pm$ 2.38    & 0.0777 $\pm$ 0.0070 \\
          & 3              & 61.54 $\pm$ 33.84     & 0.4228 $\pm$ 0.0349 \\
          & 4         & 17.50 $\pm$ 1.84    & -0.0014 $\pm$ 0.0086 \\ 
          & 5              & \textbf{8.84} $\pm$ 0.84     & \textbf{0.4575} $\pm$ 0.0402 \\          \hline
LANGUAGE   & 1              & 41.38 $\pm$ 7.10    & 0.0270 $\pm$ 0.0448 \\
          & 2               & 72.68 $\pm$ 6.33    & 0.0815 $\pm$ 0.0081 \\
          & 3              &  54.23 $\pm$ 30.18     & 0.4764 $\pm$ 0.0383 \\
          & 4          & 35.02 $\pm$ 58.49   & 0.0020 $\pm$ 0.0052 \\
          & 5              & \textbf{7.95} $\pm$ 0.87     & \textbf{0.4988} $\pm$ 0.0205 \\ \hline
MOTOR      & 1              & 117.85 $\pm$ 27.32  & -0.0016 $\pm$ 0.0456 \\
          & 2               & 77.30 $\pm$ 4.24    & 0.0782 $\pm$ 0.0110 \\
          & 3             &  57.26 $\pm$ 29.63     & 0.4156 $\pm$ 0.0548 \\
          & 4          & 21.02 $\pm$ 7.75    & 0.0023 $\pm$ 0.0090 \\ 
          & 5             & \textbf{7.80} $\pm$ 0.66     & \textbf{0.4807} $\pm$ 0.0480 \\\hline
GAMBLING   & 1              & 108.99 $\pm$ 32.83  & -0.0211 $\pm$ 0.0795 \\
          & 2               & 79.14 $\pm$ 3.65    & 0.0804 $\pm$ 0.0071 \\
          & 3             & 54.73 $\pm$  30.70    & 0.4781 $\pm$ 0.0380  \\
          & 4        & 22.63 $\pm$ 13.15   & 0.0033 $\pm$ 0.0072 \\ 
          & 5             & \textbf{7.86} $\pm$ 1.72     & \textbf{0.5014} $\pm$ 0.0301 \\\hline
SOCIAL     & 1              & 112.82 $\pm$ 36.07  & -0.0064 $\pm$ 0.076 \\
          & 2               & 79.42 $\pm$ 3.72    & 0.0772 $\pm$ 0.0088 \\
          & 3             &  47.89 $\pm$ 28.09     & 0.4912 $\pm$ 0.0353  \\
          & 4       & 18.68 $\pm$ 3.74    & 0.0025 $\pm$ 0.0071 \\ 
          & 5             & \textbf{6.81} $\pm$ 0.95     & \textbf{0.5578} $\pm$ 0.0404  \\ \hline
RELATIONAL & 1              & 147.61 $\pm$ 49.24  & -0.0265 $\pm$ 0.0754 \\
          & 2               & 81.11 $\pm$ 2.98    & 0.0706 $\pm$ 0.0081 \\
          & 3              &  54.86 $\pm$  30.68    & 0.4758 $\pm$ 0.0412   \\
          & 4        & 31.77 $\pm$ 18.30   & -0.0019 $\pm$ 0.012 \\ 
          & 5              & \textbf{8.43} $\pm$ 1.09     & \textbf{0.4858} $\pm$ 0.0460   \\\hline
          WM & 1   & 81.46 $\pm$ 29.72   & -0.0447 $\pm$ 0.0566 \\
          (Working & 2               & 77.99 $\pm$ 2.89    & 0.0799 $\pm$ 0.0109  \\
         Memory) & 3             & 56.25 $\pm$ 32.85    & 0.4767 $\pm$ 0.0579   \\
          & 4          & 103.96 $\pm$ 111.99 & -0.0041 $\pm$ 0.0089\\ 
          & 5             & \textbf{7.69} $\pm$ 1.76     & \textbf{0.4968} $\pm$ 0.0543   \\\hline  
\end{tabular}
\caption{Functional connectivity reconstruction performance of seven tasks with different models. Models are numbered as follows. 1: CGGM; 2: VAE: 3: Spectral Mapping; 4: Random $\bm{B}$; 5: CC-MRCE (Ours).}
\label{table:task-exp}
\end{table}

\begin{figure}[th]
\captionsetup[sub]{font=footnotesize}
\begin{subfigure}{.115\textwidth}
  \centering
  \includegraphics[width=.8\linewidth]{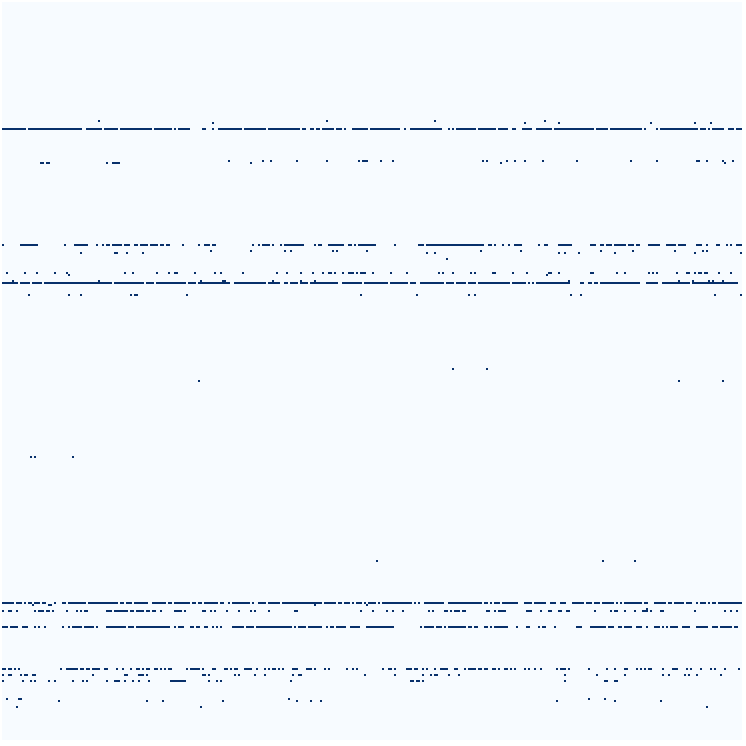}  
  \caption{EMOTION}
  \label{fig:vis_B_EMOTION}
\end{subfigure}
\begin{subfigure}{.115\textwidth}
  \centering
  \includegraphics[width=.8\linewidth]{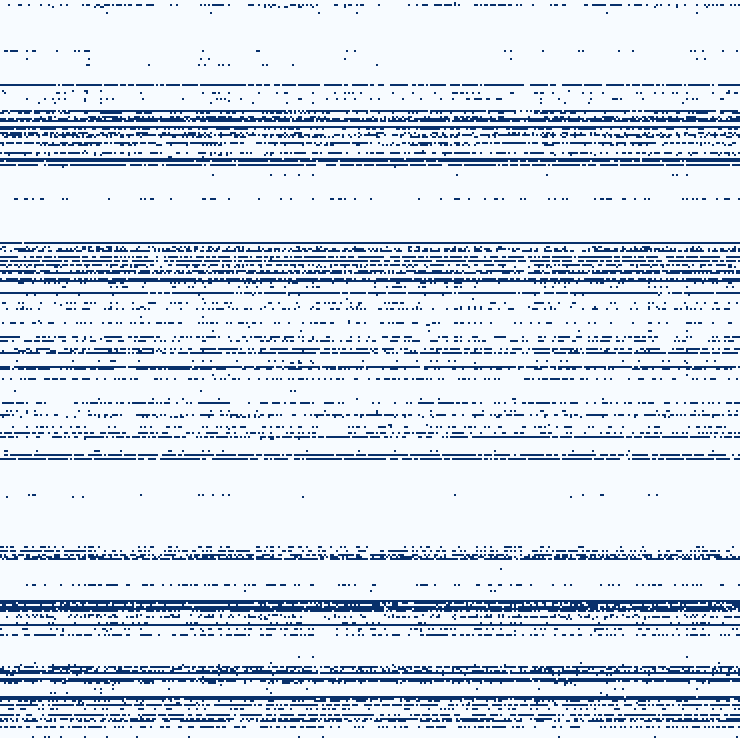}  
  \caption{LANGUAGE}
  \label{fig:vis_B_LANGUAGE}
\end{subfigure}
\begin{subfigure}{.115\textwidth}
  \centering
  \includegraphics[width=.8\linewidth]{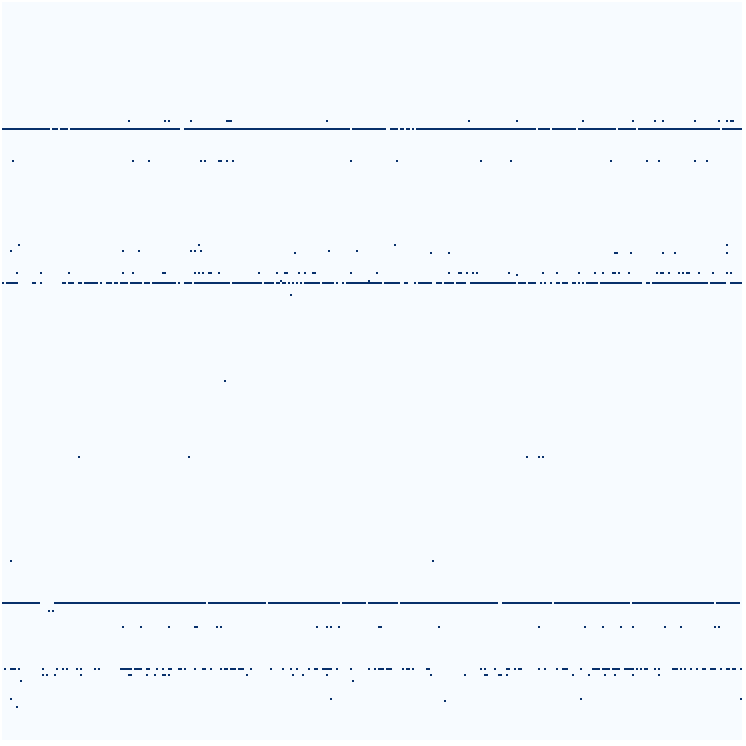}  
  \caption{MOTOR}
  \label{fig:vis_B_MOTOR}
\end{subfigure}
\begin{subfigure}{.115\textwidth}
  \centering
  \includegraphics[width=.8\linewidth]{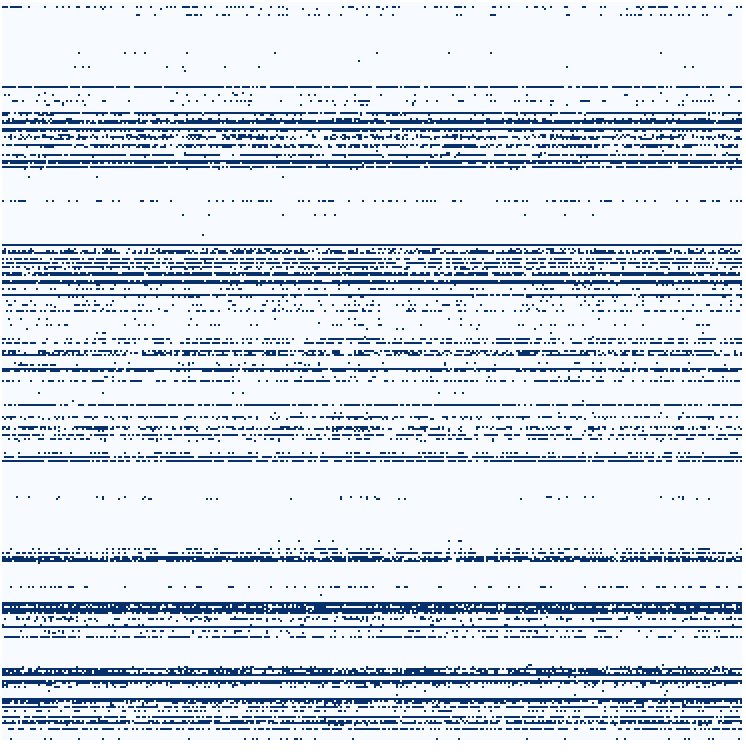}  
  \caption{GAMBLING}
  \label{fig:vis_B_GAMBLING}
\end{subfigure}
\begin{subfigure}{.115\textwidth}
  \centering
  \includegraphics[width=.8\linewidth]{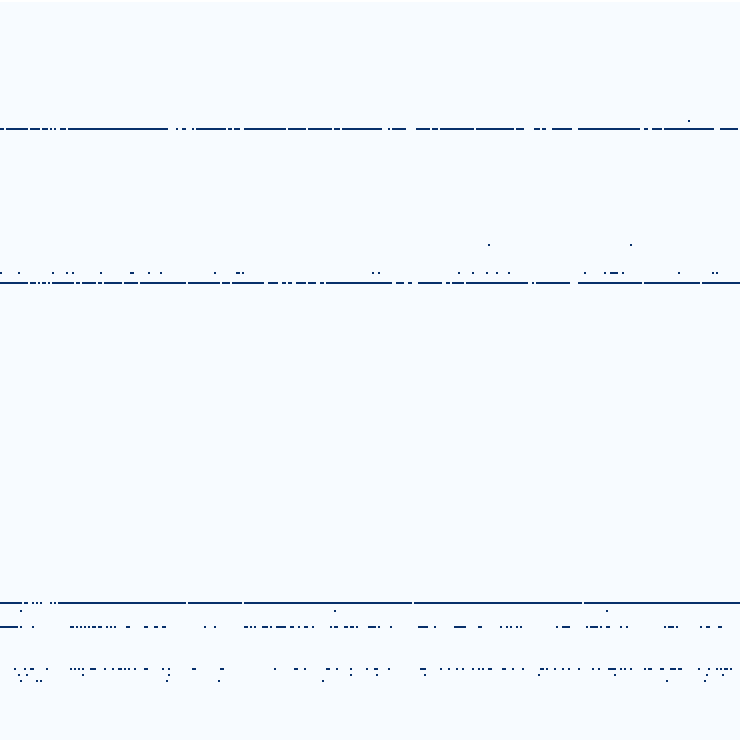}  
  \caption{SOCIAL}
  \label{fig:vis_B_SOCIAL}
\end{subfigure}
\begin{subfigure}{.115\textwidth}
  \centering
  \includegraphics[width=.8\linewidth]{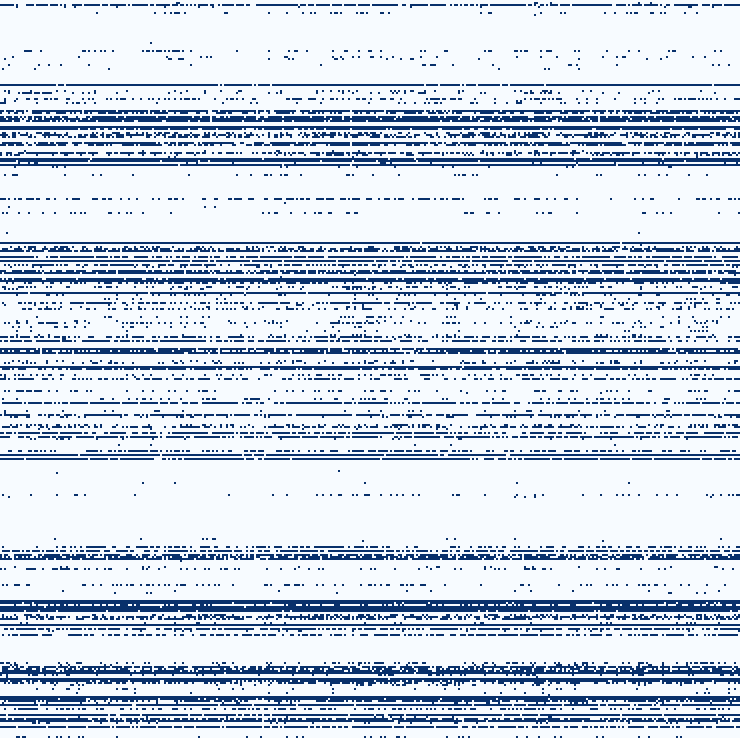}  
  \caption{RELATIONAL}
  \label{fig:vis_B_RELATIONAL}
\end{subfigure}
\begin{subfigure}{.11\textwidth}
  \centering
  \includegraphics[width=.82\linewidth]{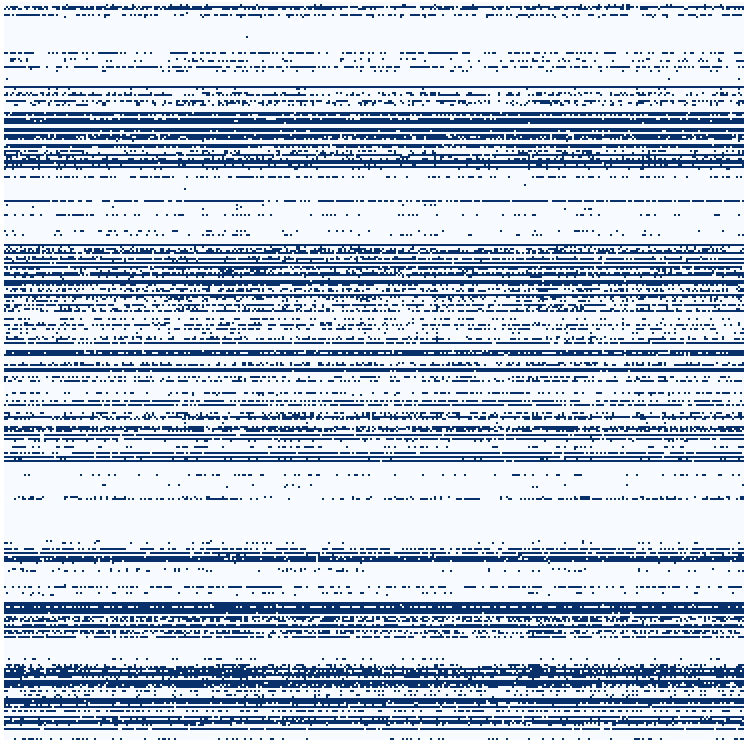}  
  \caption{WM}
  \label{fig:vis_B_WM}
\end{subfigure}
\caption[center]{Visualizations of the $\bm{B}$ matrix for seven tasks. For each task, each fold in the 10-fold cross validation may lead to different models. Here, we only show those entries that are nonzero more than five times.
}
\label{fig:vis_B}
\end{figure}


\begin{figure}[bt]
\begin{subfigure}[t]{.2\textwidth}
  \centering
  \includegraphics[width=.8\linewidth]{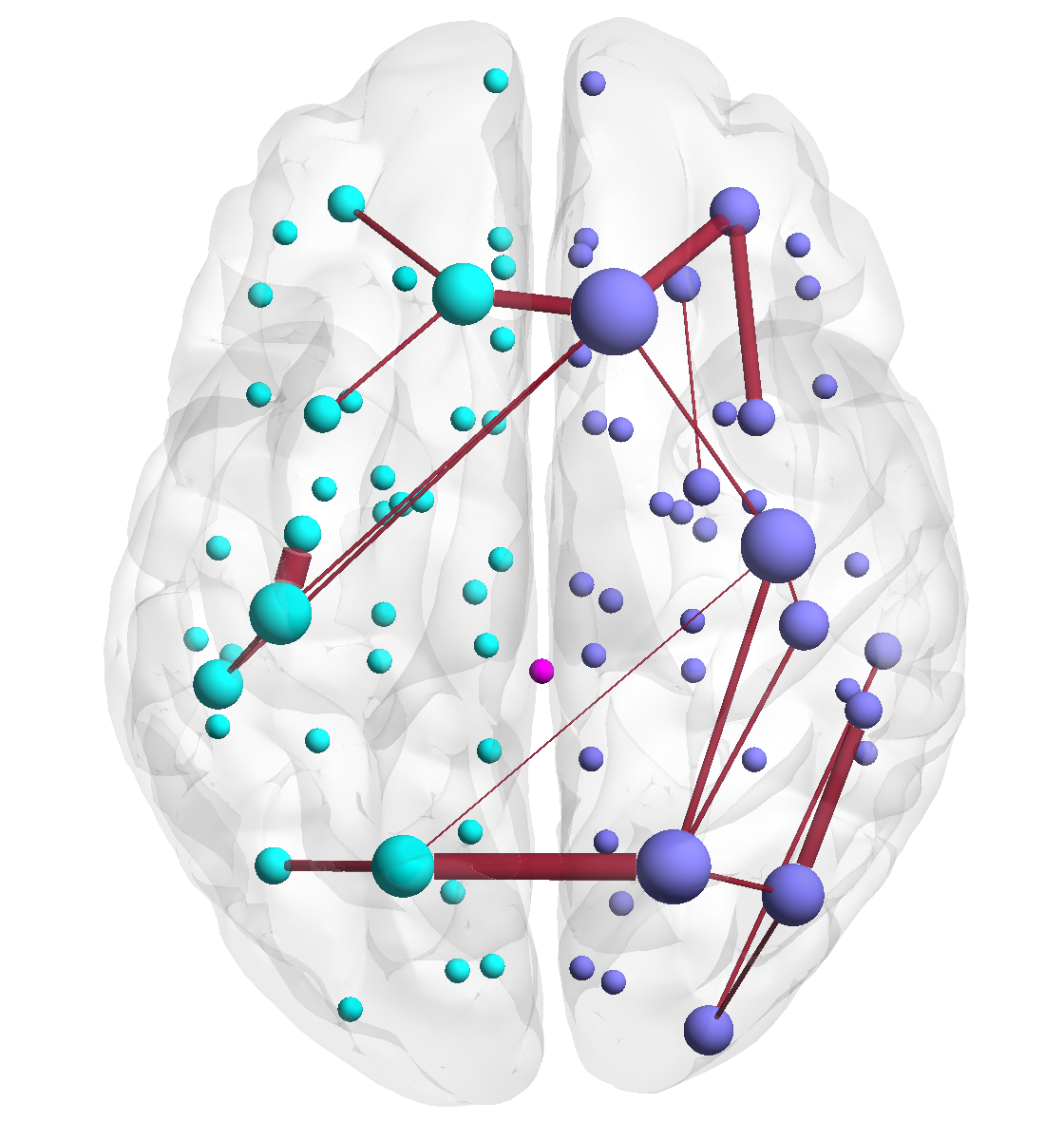}  
  \caption{Axial (from top)}
  \label{fig:brain_axial}
\end{subfigure}
\begin{subfigure}[t]{.2\textwidth}
  \centering
  \includegraphics[width=.85\linewidth]{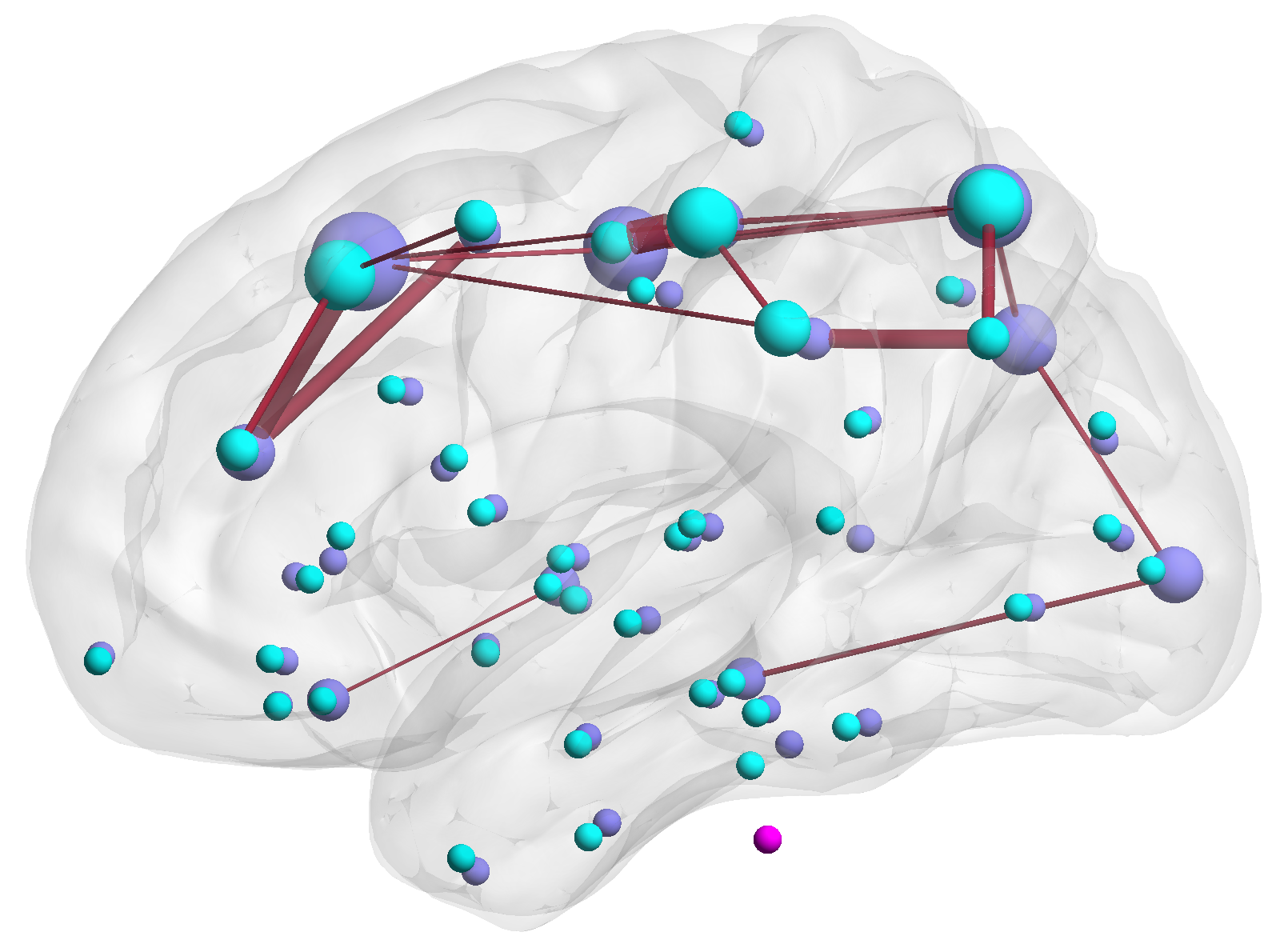}
  \caption{Sagittal (from left)}
  \label{fig:brain_sagittal}
\end{subfigure}
\centering
\begin{subfigure}[t]{.2\textwidth}
  \centering
  \includegraphics[width=.8\linewidth]{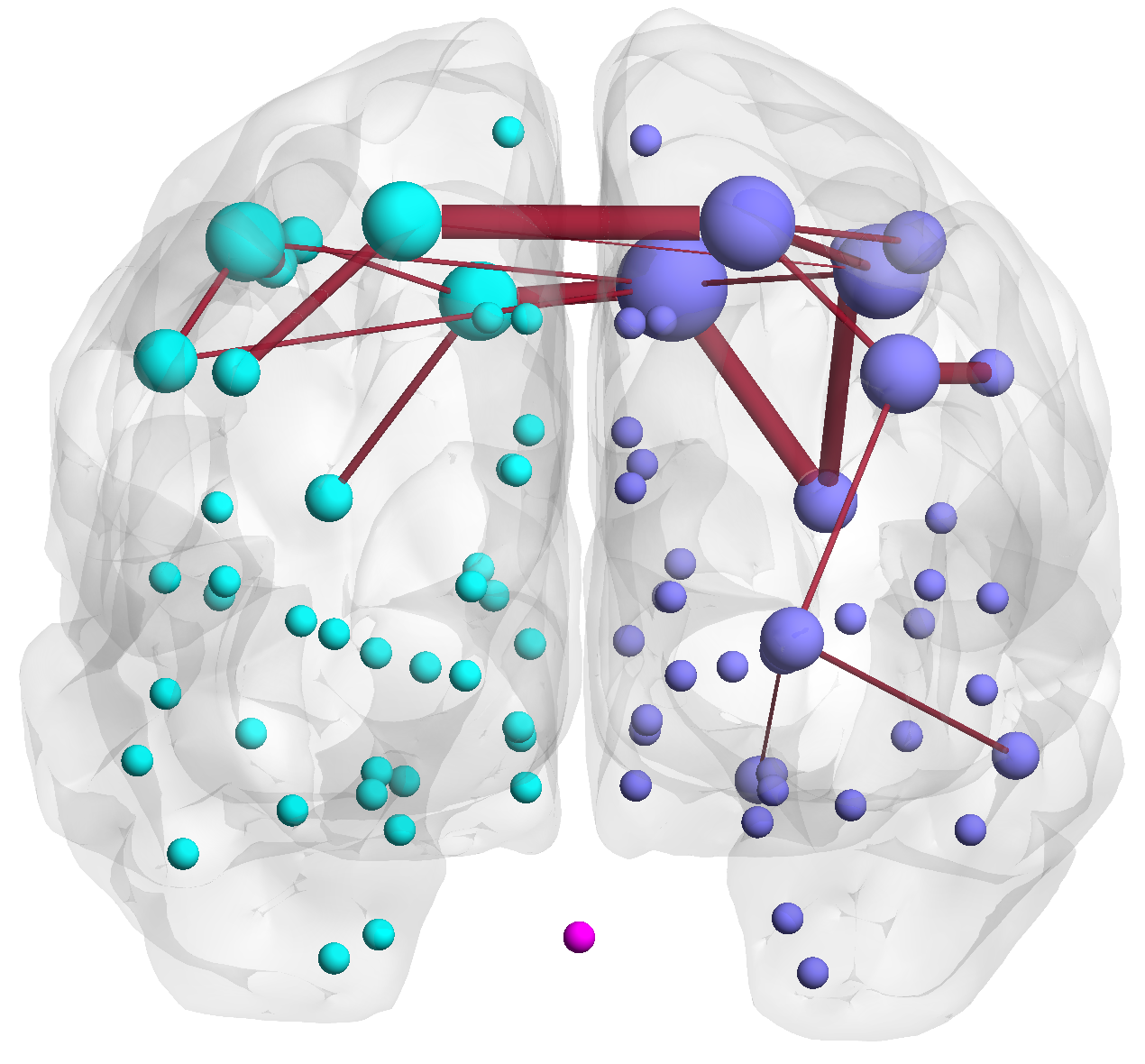} 
  \caption{Coronal (from back)}
  \label{fig:brain_coronal}
\end{subfigure}
\caption[center]{Visualization of the edges contributing to \textit{all} seven tasks. 
Node size denotes the degree and edge width denotes its importance, as in the mapping $\bm{B}$.}
\label{fig:brain_vis}
\end{figure}

\textbf{Performance Evaluation.} Regarding the ability of finding accurate mappings between SCs and FCs, we compared our model with the optimization approach CGGM, a deep learning approach Variational AutoEncoder (VAE) \cite{kingma2013auto}, and the Spectral Mapping method \cite{becker2018spectral}. For CGGM, the time required to run a 10-fold cross-validation with one set of hyperparameters on a single task ranges from one day to a week. So we ran three sets of hyperparameters on EMOTION and LANGUAGE, selected the best parameter pair, and fixed it for the rest of the five tasks. In particular, we chose penalty terms on $\bm{B}$ and $\bm{E}$ to be both 0.01. For VAE, both encoder and decoder consist of two fully connected layers, with latent variable dimensions being two. We use MSE as the training loss. In our experiments, increasing the number of layers or latent dimensions of VAE did not improve the final performance. For the Spectral Mapping method, we follow the setup of the original paper, setting maximum path length $k$ to seven.  The 10-fold cross validation results of CGGM, VAE and Spectral Mapping method are reported in Table \ref{table:task-exp}.

The assumption of Gaussian noise weakens CGGM's performance on the HCP dataset: it has unstable and large average MSE percentages across tasks, and the correlations between predictions and ground truth are very small and even negative, which are also not statistically significant with regards to $p$-values. On the other hand, VAE models have stable MSE percentage (around 80\%) and average Pearson's $r$-scores with small standard deviations across all tasks. The correlations of VAE models are weak (all around 0.08), yet statistically significant with $p$-values constantly smaller than 0.0025 for all tasks. This shows VAE learns a slightly meaningful mapping, but with such a small sample size, deep learning models are unlikely to perform well. Lastly, the Spectral Mapping method is designed for maximizing the correlation between fMRI prediction and ground truth for brain data, so it performs well as for correlation, however the prediction values are off, resulting in high MSE percentages. In all, our regression-based model performs better in both correlation and value reconstruction, showing its superiority in prediction on non-Gaussian data with small sample sizes.

\textbf{Result Interpretation.} Apart from better reconstruction performance, our model also has the advantage of result interpretability.
We can explore the SC-FC mapping through the resulting coefficient matrix $\bm{B}$s. Since our problem definition is $\text{FC} = \text{SC} \cdot \bm{B} + \bm{E}$, the $i^{th}$ row in $\bm{B}$ corresponds to $i^{th}$ edge pair in the SC vector. In the following, we say an edge $i$ exists if row $i$ of $\bm{B}$ has nonzero entries. As the experiment is run under the 10-fold cross validation setting and each training partition may generate a different mapping $\bm{B}$, 
we consider an entry in the common $\bm{B}$ to be nonzero if it is nonzero more than five times in these 10 trials. The results are shown in Figure \ref{fig:vis_B}.
From the figure, we can see for every task, several rows in $\bm{B}$ have many more nonzero entries than the others. This indicates the existence of several significant structural edges being responsible for most of the functional activities. To test if this assumption is valid, we compared $\bm{B}$s from our model to randomly generated $\bm{B}$s \textit{with the same levels of sparsity}. The resulting MSE percentages, although having large variances, often have smaller means than that of CGGM and VAE, implying the importance of sparsity level of the coupling. However, the resulting $r$-scores of predicted FCs using a random $\bm{B}$ is the lowest among all methods. Together with very large $p$-values, the results predicted by random coupling show no correlation between predictions and ground truth. This indicates that our models learned meaningful mapping information from SCs to FCs, and that \textit{structured sparsity} of $\bm{B}$ is important for getting predictions besides the level of sparsity alone. 

We now analyze the $\bm{B}$ matrices for different cognitive tasks. During fMRI data acquisitions of all seven tasks, participants are presented with visual cues, either as images or videos, and they need to use motions such as pressing buttons to complete the tasks \cite{barch2013function}. Interestingly, apart from the LANGUAGE task, the mappings learned by our model predicts the strongest contribution of left precentral and left postcentral connection, which is on the motor cortex responsible for right-side body movement. From this, we assume most participants use their right hands to conduct the required finger movements for these tasks. All mappings also contain edges in the occipital lobe, complying with the visual nature of these tasks. The visualization of common edges that exist in all seven tasks is shown in Figure \ref{fig:brain_vis}. This ``backbone'' roughly resembles the Default Mode Network \cite{greicius2009resting,raichle2015brain}.

\begin{figure}[ht]
\begin{subfigure}[t]{.23\textwidth}
  \centering
  \includegraphics[width=.9\linewidth]{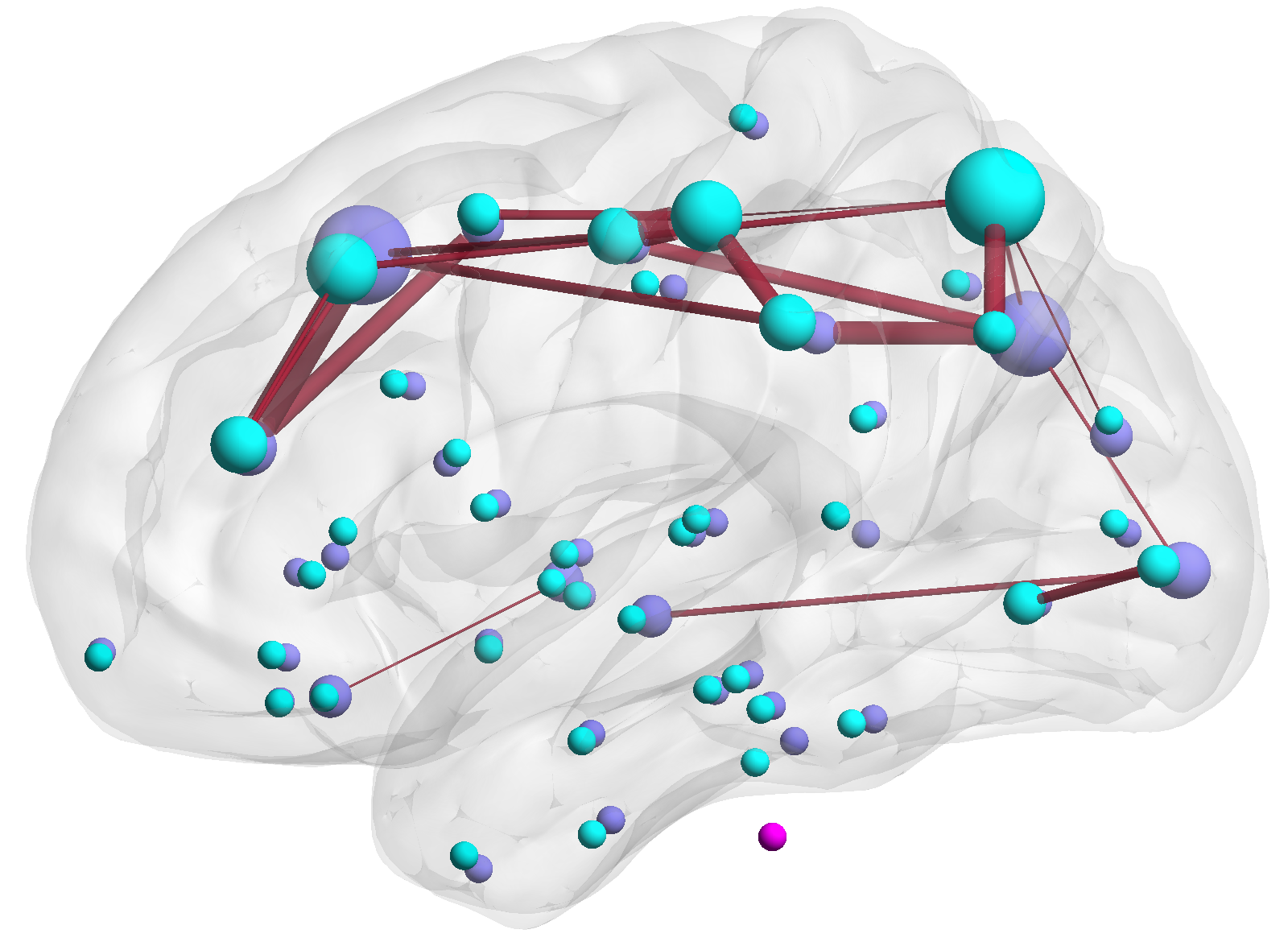}  
  \caption{LANGUAGE}
  \label{fig:brain_LANGUAGE}
\end{subfigure}
\begin{subfigure}[t]{.23\textwidth}
  \centering
  \includegraphics[width=.9\linewidth]{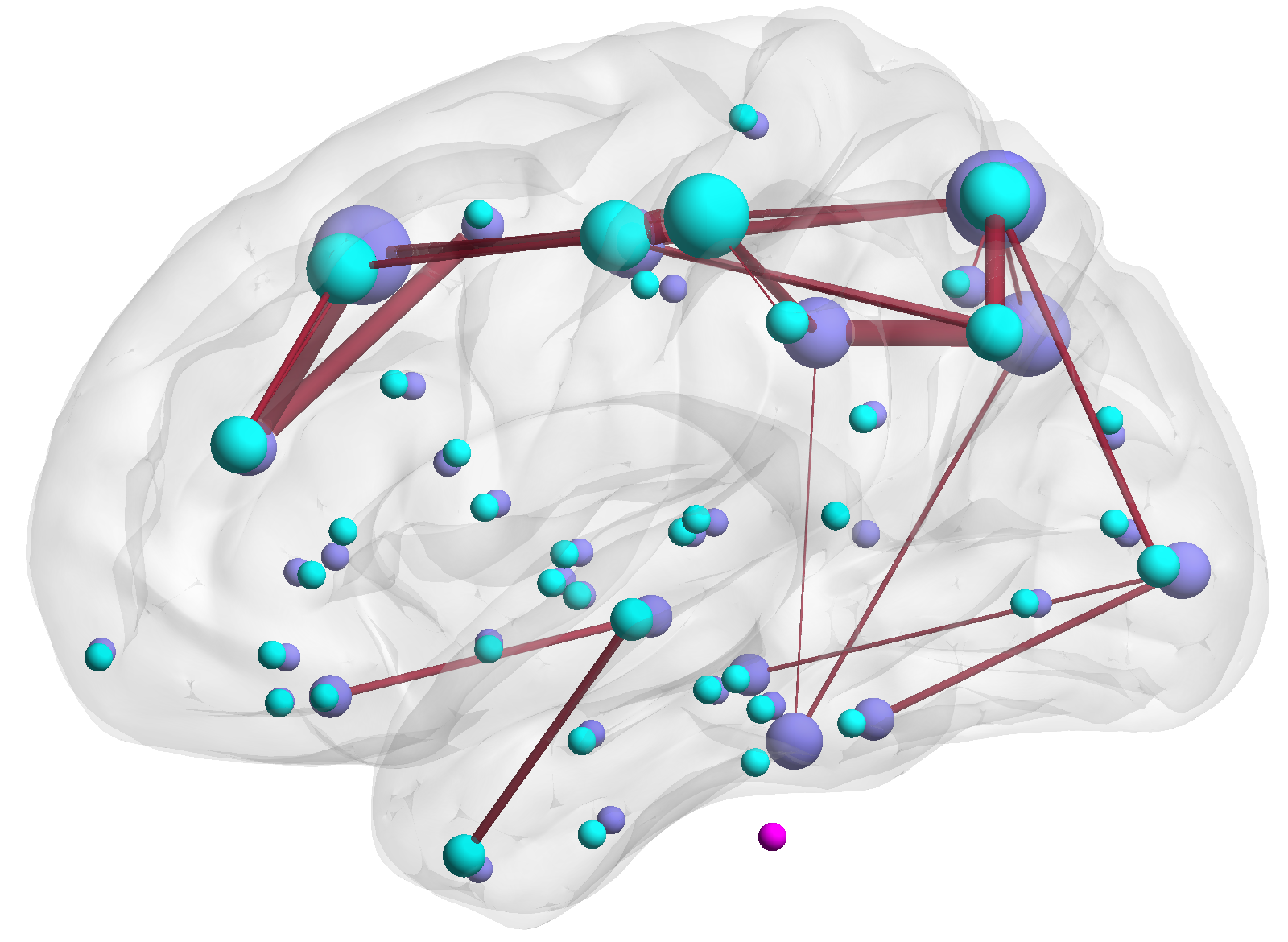} 
  \caption{GAMBLING}
  \label{fig:brain_GAMBLING}
\end{subfigure}
\begin{subfigure}[t]{.23\textwidth}
  \centering
  \includegraphics[width=.9\linewidth]{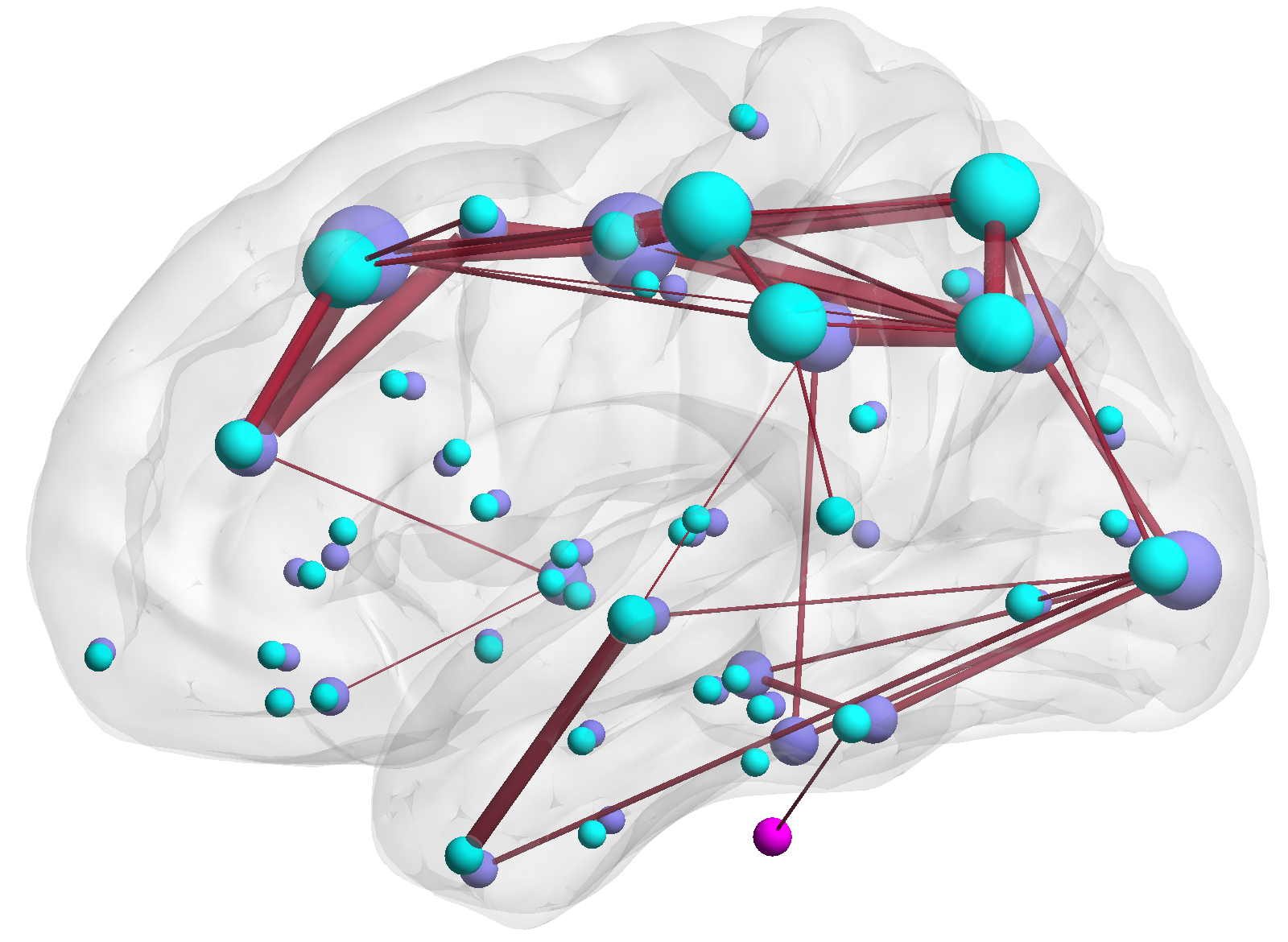}
  \caption{RELATIONAL}
  \label{fig:brain_RELATIONAL}
\end{subfigure}
\begin{subfigure}[t]{.23\textwidth}
  \centering
  \includegraphics[width=.9\linewidth]{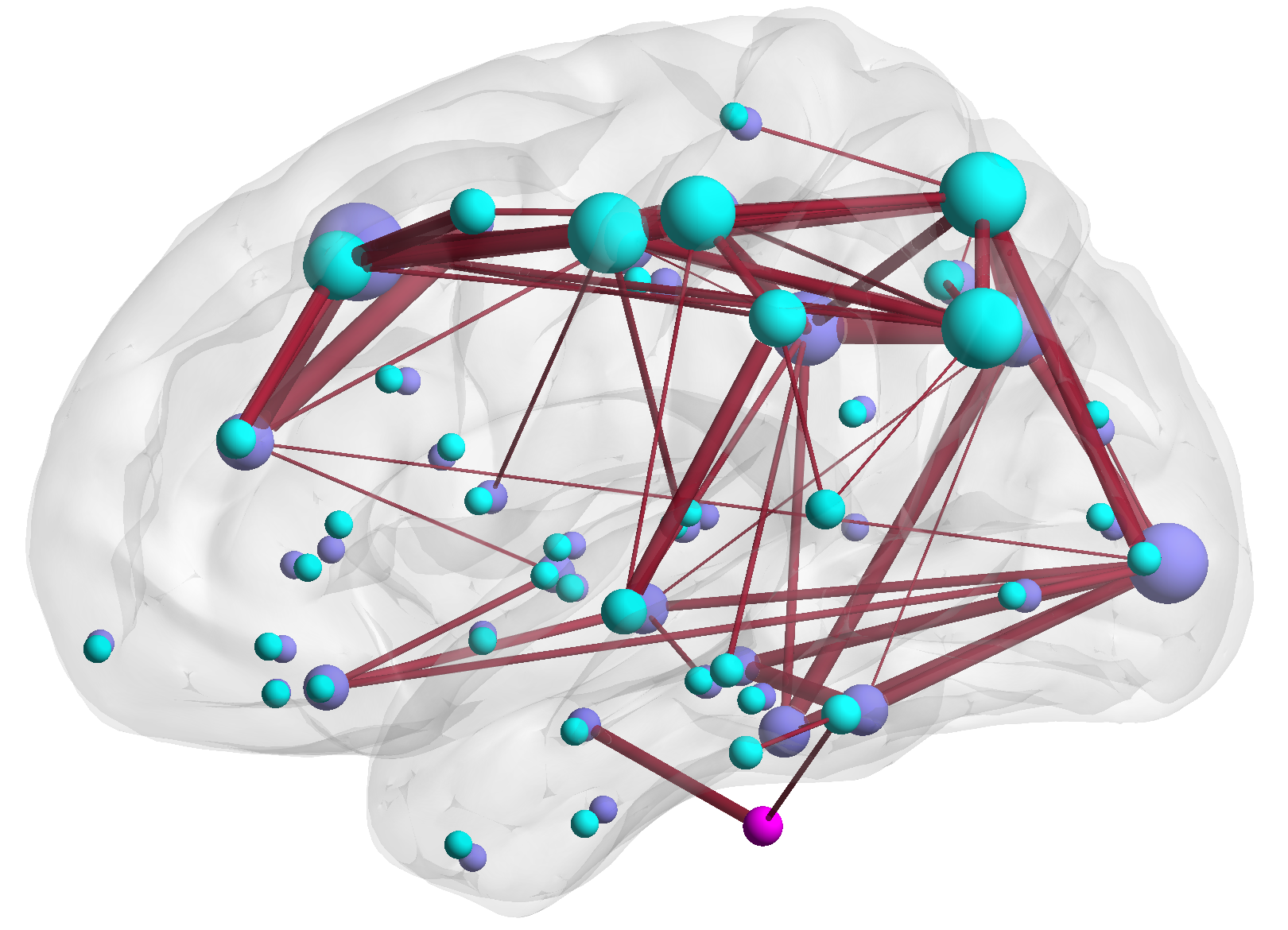}  
  \caption{WM}
  \label{fig:brain_wm}
\end{subfigure}
\caption[center]{Task-specific visualizations for high-contributing structural edges. Assuming that the maximum number of nonzero entries of a row in $\bm{B}$ is $m$, we only show the edges correspond to rows containing more than $m/2$ nonzero entries.}
\label{fig:brain_task_vis}
\end{figure}

\begin{figure}[h]
\centering
\begin{subfigure}[t]{.35\textwidth}
  \centering
  \includegraphics[width=.9\linewidth]{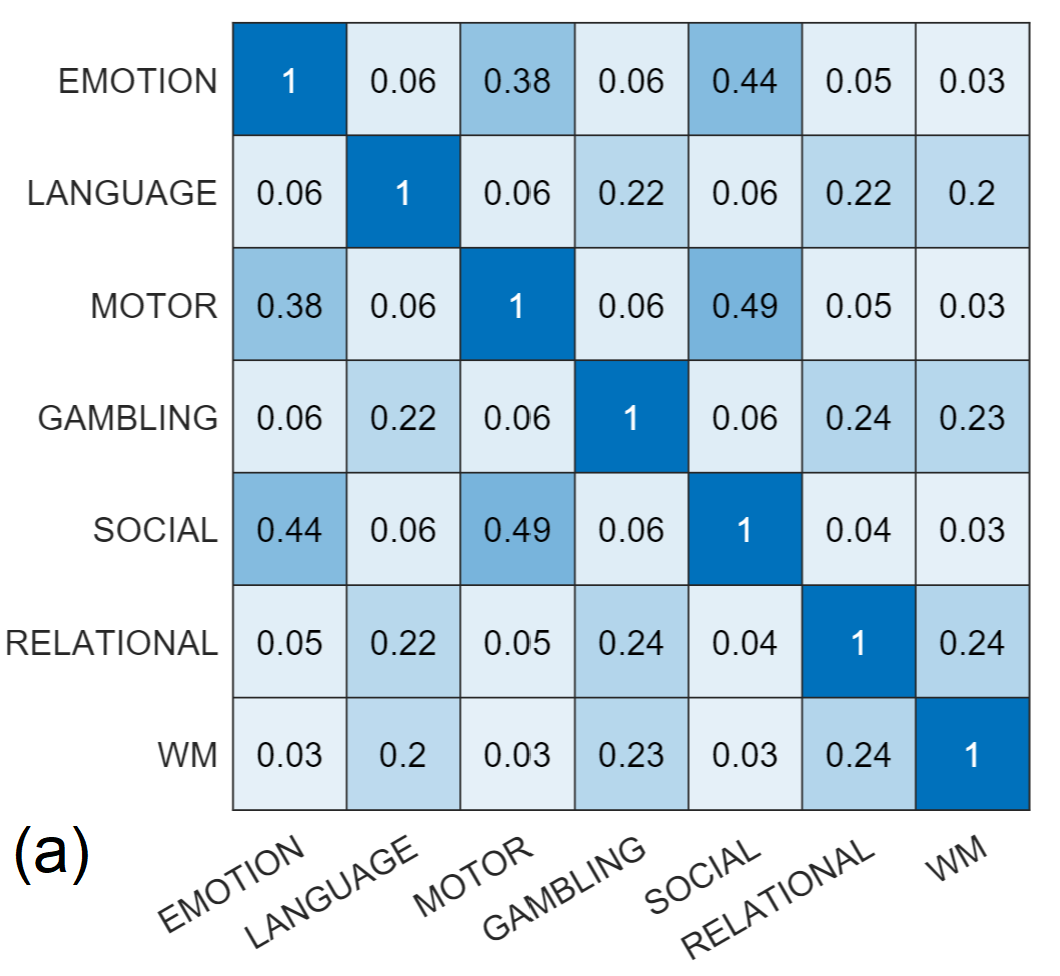}  
\end{subfigure}
\begin{subfigure}[t]{.35\textwidth}
  \centering
  \includegraphics[width=.9\linewidth]{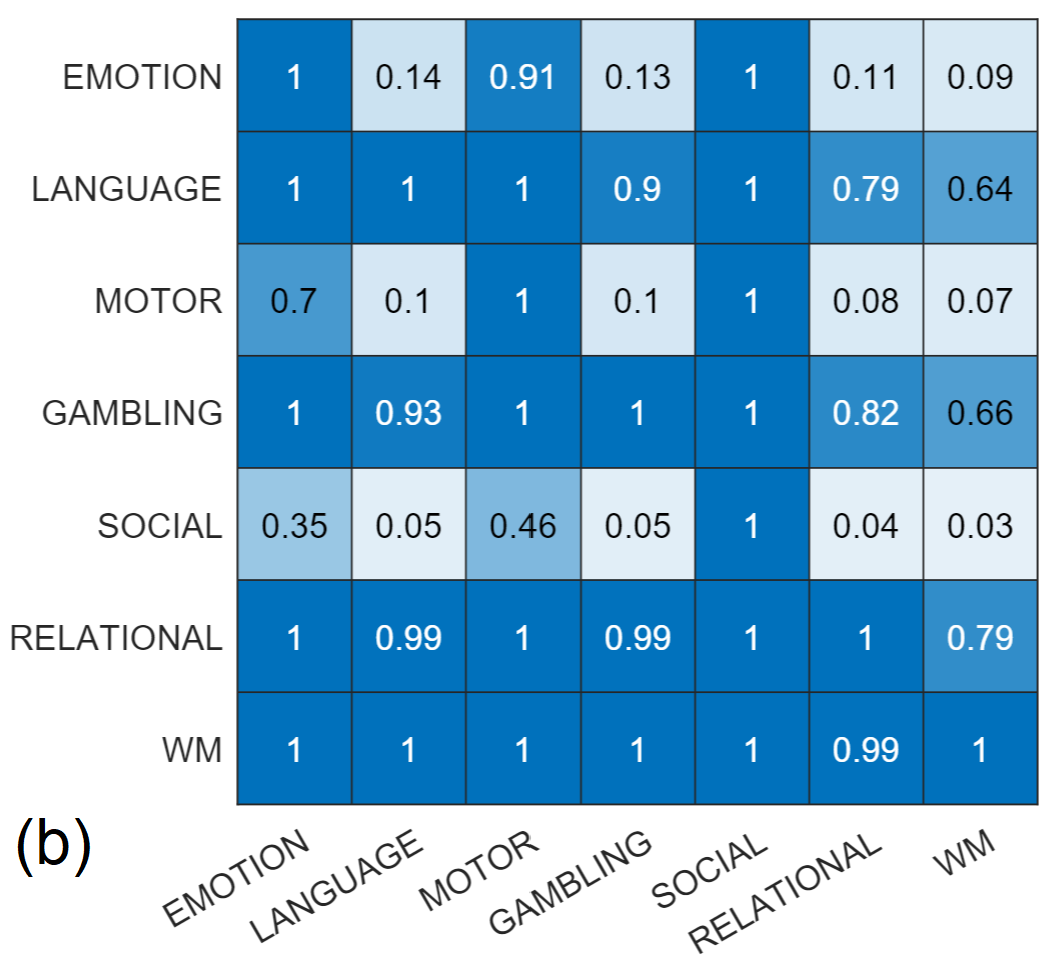}
\end{subfigure}
\caption[center]{Entry-wise and edge-wise overlap ratio for the mapping of seven tasks. 5a considers entry-wise overlap of predicted $\bm{B}$ of different tasks. The value on position (task $i$, task $j$) is the entry-wise IoU (Intersection over Union) of task $i$'s $\bm{B}$ and task $j$'s $\bm{B}$, i.e. number of nonzero entries in $\bm{B}_i \cap \bm{B}_j$ over number of nonzero entries in $\bm{B}_i \cup \bm{B}_j$. 5b considers the the SC edges responsible for different tasks. An edge is considered to exist when its corresponding row in $\bm{B}$ has nonzero entries. The value in position (task $i$, task $j$) is the number of common SC edges of task $i$ and task $j$ over the number of SC edges of task $j$.}
\label{fig:task_overlap}
\end{figure}

We then examined which structural edges contribute significantly to the functional activity under different tasks. For this, we plot the ``high-contributing'' edges in LANGUAGE, GAMBLING, RELATIONAL and WM tasks in Figure \ref{fig:brain_task_vis}. An edge is considered as high-contributing if the number of nonzero entries of its corresponding row in $\bm{B}$ is more than half of the maximum number of nonzero entries of any row in $\bm{B}$. From Figure \ref{fig:brain_task_vis}, we notice although a common backbone exists, structural connections in different brain regions are responsible for specific tasks (e.g. SCs in and around hippocampus area appear to be highly contributing to the WM FCs but not so for the LANGUAGE ones, which is consistent with the literature \cite{baddeley2011working}). We also plot both entry-wise and edge-wise overlap ratio for the mapping of seven tasks in Figure \ref{fig:task_overlap}.

Another interesting phenomenon in Figure \ref{fig:vis_B} is that the number of nonzero entries of $\bm{B}$ for LANGUAGE, GAMBLING, RELATIONAL and WM are much larger than the other three tasks: EMOTION, MOTOR and SOCIAL, although the final model for each task have a similar level of prediction performance and similar hyperparameters. This is largely caused by the nature of non-overlapping $\bm{B}$s for EMOTION, MOTOR and SOCIAL tasks: their $\bm{B}$ overlap ratios are significantly smaller than the other four tasks as shown in Table \ref{table:hcp_B_overlap}. Here we define the overlap ratio as  the number of nonzero entries in the common $\bm{B}$ (entry $ij$ being nonzero if it's nonzero more than five times) over the number of nonzero entries in $\bm{B}_{\cup} = \bm{B}_1 \cup \cdots \cup \bm{B}_{10}$ with $\bm{B}_k$ being the predicted $\bm{B}$ using the $k^{th}$ split in 10-fold cross validation. This is also why we omit these three tasks for Figure \ref{fig:brain_task_vis}, as the predicted $\bm{B}$s are not stable across the population and only a few common connections  show significant contributions. We assume this results from group heterogeneity when carrying out these tasks. Further studies with heterogeneous models for the population will be useful to verify this assumption.

\begin{table}[ht]
\small
\centering
\begin{tabular}{cccc}
\hline
EMOTION & LANGUAGE & MOTOR & GAMBLING \\ 
5.34    & 25.58    & 3.60  & 35.36
\\ \hline
SOCIAL & RELATIONAL & WM & \\
4.12   & 37.77 & 32.53 & \\\hline
\end{tabular}
\caption{Overlap ratios (\%) of predicted B (SCs-FCs mapping) across 10 folds for seven tasks.}
\label{table:hcp_B_overlap}
\end{table}

\section{Conclusions}

In this paper, we proposed a regularized regression-based approach (CC-MRCE) for jointly learning linear models and partial correlations among variables under domain constraints. Motivated by the neuroscience application of predicting functional brain activities from structural connections, the CC-MRCE method discards the Gaussian assumption and incorporates domain constraints into model estimation. We further developed a fast algorithm based on nested FISTA to solve the optimization problem. With synthetic data analysis, we demonstrated that both domain constraints and assumption of non-Gaussian data contribute to the performance improvement of CC-MRCE. Our experimental results on Human Connectome Project data show that CC-MRCE outperforms existing methods on prediction tasks and uncovers couplings that agree with existing neuroscience literature.

\section{Acknowledgments}
This project was  partially supported by funding from the National Science Foundation under grant IIS-1817046.

\section{Appendix A: Pseudocodes}

CC-MRCE (\textbf{Algorithm 2}) improves CONCORD-MRCE by imposing hard constraints on the solution space.

\begin{algorithm}[t]
\DontPrintSemicolon
\KwIn{penalty parameter $\lambda_1$ and $\lambda_2$}
\SetKwBlock{Search}{search}{end}
  Initialize $t=0$, $\hat{\bm{B}}^{(0)} = \bm{0}$ and $\hat{\bm{\Omega}}^{(0)}=\hat{\bm{\Omega}}(\hat{\bm{B}}^{(0)})$. \\
  \While{not converged}{
    {\it step 1}: Compute $\hat{\bm{S}}^{(t-1)}=\hat{\bm{S}}(\hat{\bm{B}}^{(t-1)})$ as Eq.(\ref{concord-mrce-samplecov});\\
    {\it step 2}: Update $\hat{\bm{\Omega}}^{(t)}=\hat{\bm{\Omega}}(\hat{\bm{S}}^{(t-1)})$ as Eq. (\ref{concord-mrce-EM-omega}) by calling CONCORD($\hat{\bm{S}}^{(t-1)}$) ;\\
    {\it step 3}: Update $\hat{\bm{B}}^{(t)}=\hat{\bm{B}}(\hat{\bm{\Omega}}^{(t)})$ as Eq.(\ref{concord-mrce-EM-beta});\\
  }
  \Return{$\bm{B}$ and $\bm{\Omega}$}
\caption{CONCORD-MRCE}\label{algorithm-concord-mrce}
\end{algorithm}

\begin{algorithm}[t]
\DontPrintSemicolon
\KwIn{penalty parameter $\lambda_1$ and $\lambda_2$, convex constraint set $C$.}
\SetKwBlock{Search}{search}{end}
  Initialize $\hat{\bm{B}}^{(0)} = \bm{0}$ and $\hat{\bm{\Omega}}^{(0)}=\hat{\bm{\Omega}}(\hat{\bm{B}}^{(0)})$. \\
  \While{not converged}{
    {\it step 1}: Compute $\hat{\bm{S}}^{(t-1)}=\hat{\bm{S}}(\hat{\bm{B}}^{(t-1)})$ as Eq.(\ref{concord-mrce-samplecov});\\
    {\it step 2}: Update $\hat{\bm{\Omega}}^{(t)}=\hat{\bm{\Omega}}(\hat{\bm{S}}^{(t-1)})$ by calling Constrained-CONCORD($\hat{\bm{S}}^{(t-1)}$,$E$,$\lambda_1$);\\
    {\it step 3}: Update $\hat{\bm{B}}^{(t)}=\hat{\bm{B}}(\hat{\bm{\Omega}}^{(t)})$ as Eq.(\ref{concord-mrce-EM-beta});\\
  }
  \Return{$\bm{B}$ and $\bm{\Omega}$}
\caption{CC-MRCE}\label{algorithm-cc-mrce}
\end{algorithm}

 In every step of the while loop, CC-MRCE calls sub-function Constrained-CONCORD to estimate the partial correlation matrix $\bm{\Omega}^{(t)}$ under hard constraints, i.e. the solution $\bm{\Omega}^{(t)}$ shares the same zero and nonzero pattern as matrix $E$ (shown in \textbf{Algorithm 3}).  Such constraints can be replaced by any set of convex constraints on $\bm{\Omega}^{(t)}$. Note that superscript $t'$ and $t$ in \textbf{Algorithm 3} are iteration counters of inner and outer stages in the Constrained-CONCORD algorithm, respectively.

\begin{algorithm}[h]
\DontPrintSemicolon
\KwIn{sample covariance matrix $\bm{S}$, sparsity pattern $E$, penalty parameter $\lambda_2$}
\KwOut{partial correlation matrix $\bm{\Omega}$}
\SetKwBlock{Search}{search}{end}
  set $\bm{\Theta}^{(1)}=\bm{\Omega}^{(0)}\in \mathbb{S}_M^p$, $\alpha_1 = 1$, $\tau_{(0,0)} \leq 1$, $c < 1$ \;
  \While{not converged}  {
    $\bm{G}^{(t)} = \nabla h_1(\bm{\Theta}^{(t)})$\;
    \Search(largest $\tau_t \in \lbrace c^j \tau_{(t,0)} \rbrace_{j=0,1,\cdots}$){
        $\bm{A}^{(t)} = \bm{\Theta}^{(t)} - (n\tau_t/2) \bm{G}^{(t)}$ \;
        set $\tilde{\bm{\Theta}}^{(1)}=\bm{H}^{(0)}\in \mathbb{S}_{d\times d}$, $\tilde{\alpha}_1 = 1$, $\kappa_{(0,0)} \leq 1$, $\tilde{c} < 1$\;
        \While{not converged}{
            $\tilde{\bm{G}}_X^{(t')} = \nabla g_1(\bm{H}_X^{(t')})$\;
            \Search(largest $\kappa_{t'} \in \lbrace \tilde{c}^{j'} \kappa_{(t',0)} \rbrace_{j'=0,1,\cdots}$){
                $ \bm{H}_X^{(t')} = \operatorname{prox}_{g_2}\left(\tilde{\bm{\Theta}}_X^{(t')} - \kappa_{t'}\tilde{\bm{G}}_X^{(t')}\right)$ \;
                check backtrack line search criterion \;
            }
            $\tilde{\alpha}_{t'+1} = (1+\sqrt{1+4\tilde{\alpha}_{t'}^2})/2$ \;
            $\tilde{\bm{\Theta}}^{(t'+1)} = \bm{H}_X^{(t')} + \frac{\alpha_{t'}-1}{\alpha_{t'+1}}
            \left(\bm{H}_X^{(t')} - \bm{H}_X^{(t'-1)}\right)$ \;
            compute $\kappa_{(t'+1,0)}$
        }
        $\bm{\Omega}^{(t)}  = \mathbb{P}_C({\bm{A}_X}^{(t)}-\gamma \lambda_2 {\bm{H}_X}^{(t)}) + \bm{A}_D^{(t)} $ \;
        check backtrack line search criterion \;
    }
    $\alpha_{t+1} = (1+\sqrt{1+4\alpha_t^2})/2$ \;
    $\bm{\Theta}^{(t+1)} = \bm{\Omega}^{(t)} + \frac{\alpha_{t}-1}{\alpha_{t+1}}(\bm{\Omega}^{(t)} - \bm{\Omega}^{(t-1)})$ \;
    compute $\tau_{(t+1,0)}$
  }\label{endwhile}
\caption{Constrained-CONCORD}\label{algorithm-nc-concord}
\end{algorithm}

\begin{lemma} Let $L(g_1)$ be the Lipschitze constant of the gradient of objective function $g_1(\bm{H}_X)$, then $L(g_1) \leq 2 \lambda_2^2 \gamma^2$.
\end{lemma}

\noindent
Although we use line-search to pick a proper step length $\kappa_{t'}$ in the inner-loop of \textbf{Algorithm 3}, it can be replaced with a constant step length $\kappa_{t'} = 2 \lambda_2^2 \gamma^2 $ according to the above lemma.

\bibliography{reference-2021.bib}

\end{document}